# The Impact of Processing Parameters on High-Accuracy Measurements in UAV Photogrammetry

Paweł Ćwiąkała[1,*], Edyta Puniach[1], Elżbieta Pastucha[2], Wojciech Gruszczyński[1]

[1] AGH University of Krakow, al. A. Mickiewicza 30, 30-059 Krakow, Poland

[2] The Mærsk Mc-Kinney Møller Institute, University of Southern Denmark, Campusvej 55, DK-5230 Odense, Denmark

* e-mail: pawelcwi@agh.edu.pl

**Abstract**

Unmanned aerial vehicle (UAV) photogrammetry is increasingly used in applications requiring high accuracy, such as determining ground surface changes caused by landslides, mining, or microrelief transformation. While acquisition strategies have been widely studied, the influence of the processing workflow—particularly Bundle Block Adjustment parameter settings—remains insufficiently explored. This study addresses this gap through a systematic, full-factorial evaluation of 768 processing variants applied to ten UAV datasets collected over 1.5 years in a 220 ha study area. Eight key parameters were analysed. The results show substantial variability in final 3D accuracy: the best-performing variant achieved a root mean square error (RMSE) of 16 mm, whereas the weakest reached 303 mm. The most influential factors were the number of ground control points, the application of additional camera calibration corrections, and the use of the Post-Processing Kinematic GNSS method for determining camera projection center coordinates. The study also evaluates how workflow optimization affects the accuracy of displacement, tilt changes, and horizontal strain determination. While random displacement errors remained stable (RMSE of ~6–7 mm), systematic errors were significantly reduced by over half in all axes, with vertical median absolute error decreasing from 14 mm to 7 mm in the optimized configuration compared to the baseline previously used by the authors. This study provides the first large-scale, practice-oriented assessment of how processing parameter selection shapes the accuracy of both photogrammetric products and deformation indices determination. The results offer actionable guidance for developing more robust and repeatable UAV photogrammetry workflows tailored to high-precision monitoring.

## 1. Introduction

Unmanned Aerial Vehicle (UAV) photogrammetry has become an essential tool for surveyors, engineers, and scientists, facilitated by the widespread availability of cost-effective, user-friendly UAV systems equipped with non-metric RGB cameras. These platforms provide a compelling alternative for mapping small- to medium-sized areas at centimeter-level resolution. Concurrently, the market for UAV data processing software has rapidly evolved, emphasizing automation to minimize user intervention. A prominent trend in commercial photogrammetric software is the move towards "black-box" solutions, which significantly limit the user's ability to adjust processing parameters manually. While default settings often yield adequate accuracy for many applications, especially with high-quality input data, this approach may prove insufficient when data quality is suboptimal or when exceptionally high accuracy is demanded. In such cases, relying solely on automated workflows can lead to unsatisfactory results. Consequently, a deeper understanding of the influence of Bundle Block Adjustment (BBA) parameters and other processing parameters is critical for improving the accuracy and reliability of UAV photogrammetry products.

Research in UAV photogrammetry has largely focused on optimizing data acquisition strategies to enhance accuracy. Numerous studies have explored how various acquisition parameters—such as UAV positioning method, flight altitude, terrain-following mode, flight pattern, image overlap, and camera tilt angle—affect BBA outcomes [1–6]. In parallel, the number and spatial distribution of ground control points (GCPs) have been extensively investigated [1, 7–16]. Commonly small UAVs with integrated cameras, such as the DJI Phantom 4 [7–9; 11–16] or the DJI Mavic 3 [17], are utilized in studies on UAV photogrammetry accuracy, although other platforms featuring cameras like the SONY a6000 or the Sony RX1RII are also employed [1].

UAV positioning methods have evolved significantly, shifting from indirect georeferencing—heavily reliant on numerous GCPs [18–20]—to direct georeferencing based on high-accuracy global navigation satellite system (GNSS)-based techniques like Post-Processing Kinematic (PPK), Real-Time Kinematic (RTK), and Real-Time Network (RTN). Many studies have compared the impact of these different positioning methods on BBA accuracy [1, 7, 13, 15, 16, 21, 22]. For instance, Pargieła and Rzonca [1] have compared autonomous GNSS positioning with PPK GNSS using a local base station, while Forlani et al. [7] evaluated multiple methods, including RTN, RTK, PPK, and autonomous GNSS. The influence of the GNSS base station's location has also been scrutinized; a study by Miron et al. [15] found no significant accuracy differences when using various continuous operating reference stations (CORS) up to 2.5 km away, while McMahon et al. [21] demonstrated that for greater distances to CORS (about 8–22 km) PPK GNSS direct georeferencing with a local base station offers notable advantages over using CORS. That study also found that errors decrease when the base station is closer to the survey area or operates at a higher sampling rate.

Although some research suggests that GCPs are no longer necessary when using high-accuracy direct georeferencing [1, 13, 16, 21, 23, 24], onboard GNSS receivers are susceptible to errors [12, 25–27]. Therefore, the integration of direct georeferencing with a limited number of well-distributed GCPs remains a widely recommended practice to ensure the accuracy and reliability of results [10, 11, 28, 29].

A review of the literature in this field reveals several common features and limitations. The published studies consistently report centimeter-level accuracies in the final UAV photogrammetry products. Root mean square errors (RMSE) at check points (CPs) typically range from 1 to 3 cm in the horizontal components and from 2 to 5 cm in the vertical component [1, 7–11, 13, 15, 16], though higher values are occasionally reported [12]. Ground sampling distances (GSDs) reported in these studies typically range from 17 to 35 mm. One notable exception is presented by Nesbit et al. [11], where the GSD ranged from 7 to 17 mm; however, this case involved a very small study area of only 0.4 ha. Most studies are conducted over similarly small areas—generally less than 10 ha [1–3, 7–9, 11, 13, 15, 30]—although a few have investigated significantly larger sites [10, 14]. The majority are based on data from a single measurement campaign, with only a few exceptions, where the number of measurement series remains low (up to three) [8, 12]. A key focus across studies is the impact of GCP quantity on BBA accuracy. For instance, Dai et al. [2, 3], in an extensive analysis of 177,000 scenarios, demonstrated that GCPs have the strongest effect on accuracy: even a few well-distributed points significantly improve results, while adding more further reduces spatial error correlation. The number of GCPs tested varies widely, from none [1, 10, 11, 13, 15, 16] to several dozen [8]. For example, Zeybek [8] used 30 GCPs over a 2.4 ha area—equivalent to 12.5 GCP/ha. Similarly, the number of CPs ranges from none [11] to more than 60 [1]. Importantly, many studies do not use a fixed CP set; instead, CPs and GCPs are often selected interchangeably depending on the scenario being tested [1, 7, 10, 11, 13, 16]. In other words, the same points serve as both GCPs and CPs across different configurations. GCP and CP positions are almost always measured using RTK/RTN GNSS or total stations.

Another important issue for the results of BBA is the camera's interior orientation parameters, which are typically derived through self-calibration during the BBA [31–33]. Recent advancements center on optimizing self-calibration by strengthening the image network geometry through the integration of oblique imagery [2, 3, 5, 6], multi-altitude flight plans [1–3, 7–9, 11, 13, 15], and terrain-following modes [6].

Traditionally, UAV photogrammetry has relied on the Brown distortion model to estimate camera calibration parameters [34]. Some research has investigated how the selection of specific calibration parameters, particularly in conjunction with the number of GCPs, influences the overall accuracy of the BBA [35, 36]. More advanced models, such as the 96-parameter model proposed by Tang et al. [37], have also been introduced and are now widely adopted, largely due to their implementation in standard software packages like Agisoft Metashape. Alternative calibration strategies for UAV-mounted cameras are also being explored, though they remain rarely adopted in practice [38–40].

The software used to perform BBA is also crucial, as the available parameters and their value ranges that the user can set vary across platforms. Among data processing software, two commercial packages are dominant: Agisoft Metashape [1, 7, 10, 13, 15, 16, 18, 20] and Pix4D [8, 9, 11, 12, 14, 19]. Performance of these packages, together with Inpho UAS Master, Context Capture, and MicMac, was assessed by Casella et al. [41]. According to the research results, Agisoft Photoscan, UAS Master, and Pix4D consistently produced good results, while Context Capture and MicMac showed good results for the horizontal plane but exhibited large residuals for elevation. Results presented by Kostrzewa et al. [42], who compared the accuracy and efficiency of OpenDroneMap (ODM) and Agisoft Metashape, demonstrated that the open-source solution could achieve comparable camera calibration results to the commercial one. However, the RMSE on CPs in ODM was approximately twice as high as in Agisoft Metashape.

Despite the central role of BBA in determining the accuracy of UAV photogrammetry products, most articles do not provide detailed information about processing parameters. Such details are disclosed only sporadically [7, 41, 42]. Although general recommendations can be found in user manuals or institutional guidelines [43], no study to date has systematically evaluated the importance of individual processing parameters or assessed the consistency of their impact across diverse datasets. This gap was highlighted in a study by Puniach et al. [44], who identified systematic errors in high-resolution displacement analyses likely originating from the UAV data processing, even with intuitively reasonable parameter settings. Motivated by these findings, this study systematically investigates how different processing parameter configurations affect the accuracy of UAV photogrammetry products. The goal is to identify parameter combinations that consistently yield high-accuracy results, regardless of acquisition conditions. The analysis leverages a large and diverse dataset from ten independent surveys over a 220 ha area, collected over 1.5 years under varying environmental conditions. Data from each survey were processed using 768 unique parameter combinations (processing variants), and accuracy was assessed against an independent and constant set of CPs. By identifying the most influential parameters, this research aims to establish more reliable processing workflows and enhance the accuracy of derived products, particularly for monitoring ground surface deformation.

Given the increasing use of UAV photogrammetry to detect ground surface changes—such as terrain displacement caused by mass movements [45–47] or underground mining [44, 48–50], or the evolution of landforms and microrelief [51, 52]—the findings presented in this study may contribute to improving the accuracy of such analyses. Until now, analyses of this kind have typically been conducted with accuracies on the order of a few centimeters, with the highest levels of accuracy reported by Puniach et al. [44].

To summarise, the principal contributions of this study are as follows:

(1) a full-factorial evaluation of 768 processing variants applied to ten UAV datasets collected over 1.5 years across a 220 ha area;
(2) identification of the parameters that most significantly affect the accuracy of photogrammetric products and ground surface deformation indices determination;
(3) assessment of both systematic and random error components in deformation indices associated with processing variant selection;
(4) formulation of practical guidelines for designing robust, repeatable, and accuracy-oriented UAV photogrammetry workflows for deformation monitoring.

All conclusions refer to the specific measurement conditions under which the analyses were performed. Under different conditions—particularly those related to study area size, image resolution, or the accuracy of GCP coordinate determination—the relationships between photogrammetry product accuracy and the adopted processing parameter values may differ.

The remainder of this paper is structured as follows. Section 2 describes the study area, data acquisition, the tested parameters, and the methodologies for evaluation. Section 3 presents the results of the accuracy analysis and identifies the best-performing variant. Section 4 discusses the findings. Finally, Section 5 provides the research summary.

## 2. Materials and methods

The study area is introduced at the beginning of this section. This is followed by a description of the UAV photogrammetry data acquisition and its processing across 768 variants, along with a method of evaluation of how the processing parameters influence accuracy. The section also details the workflow for determining and assessing the accuracy of UAV-derived ground surface displacements, tilt changes, and horizontal strains.

### 2.1. Study data

The study area, spanning approximately 220 ha, is situated in the village of Jerzmanowice (50°12′ N, 19°45′ E), within the Lesser Poland Voivodeship, Poland. This area is not affected by any known factors that would induce ground surface displacement. The topography is relatively uniform (Fig. 1), with local inclines reaching up to 17 degrees; however, the majority of the area exhibits a slope of 5 degrees or less. The land cover is predominantly composed of fields and meadows with sparse tree growth. Buildings, consisting mainly of single-family homes and agricultural structures, are concentrated along roads that delineate the site's boundaries and traverse its central part.

Periodic UAV photogrammetry surveys were performed over the study area. A total of 10 consecutive series were analyzed, conducted in May 2022, June 2022, October 2022, December 2022, January 2023, March 2023, April 2023, July 2023, August 2023, and September 2023. The surveys were carried out using a DJI Matrice M300 (DJI, Shenzhen, China) equipped with a DJI Zenmuse P1 camera and a native DJI DL 35mm F2.8 LS ASPH lens. The camera featured a full-frame CMOS sensor (35.9 mm × 24.0 mm) with a resolution of 8192 × 5460 pixels (45 Mpx). The DJI Matrice M300 multicopter is equipped with a Global Navigation Satellite System (GNSS) Real-Time Kinematic (RTK) module, enabling precise determination of the device's trajectory. The photogrammetric mission plan was identical for all series. Flight stripe spacing and UAV speed were configured to ensure 80% forward and 60% side image overlap, and the flight altitude was set to achieve a ground sampling distance of 10 mm (terrain follow mode was applied). Approximately 7,500 images were acquired during each session.

A permanent network of GCPs and CPs was established within the study area (Fig. 1). The two sets were defined as disjoint to ensure independent accuracy assessment across all processing variants. All GCPs and CPs were measured once, and the same coordinates were used consistently in every processing variant. Prior to each measurement series, the coordinates of all network points were rechecked to confirm their stability.

The coordinates of the base station, as well as the GCPs and CPs located along the roads, were determined using the static GNSS method with eight Leica GS16 receivers mounted on tripods. Observations were referenced to continuously operating reference stations (CORS) of the nationwide ASG-EUPOS (Active Geodetic Network – European Position Determination System) network. Specifically, stations KRA1, PROS, and KATO—located approximately 20 km, 40 km and 50 km from the study area, respectively—were used. Static observations were recorded at a 5-second sampling rate during 45-minute sessions at each point. During each session, simultaneous measurements were performed at the base station and at seven additional points, with at least two points reoccupied from the previous session to ensure network consistency. Observations at the base station were continuous throughout the entire network survey and lasted approximately 7 hours. Post-processing of the GNSS vectors yielded coordinate accuracies (RMSE) of approximately ±1.0 cm horizontally and ±1.5 cm vertically for the base station and the photogrammetric network points measured in static mode.

The remaining GCPs and CPs were measured using the GNSS RTK method with Leica GS16 receivers, referencing the base station situated near the center of the study area (with coordinates determined from the static survey described above). For these measurements, the rover was mounted on a 2-meter handheld pole, and readings were taken in 30-second epochs. Based on the equipment specifications, the coordinates of these points were estimated with an accuracy of approximately ±1–2 cm horizontally and ±2–3 cm vertically relative to the base station.

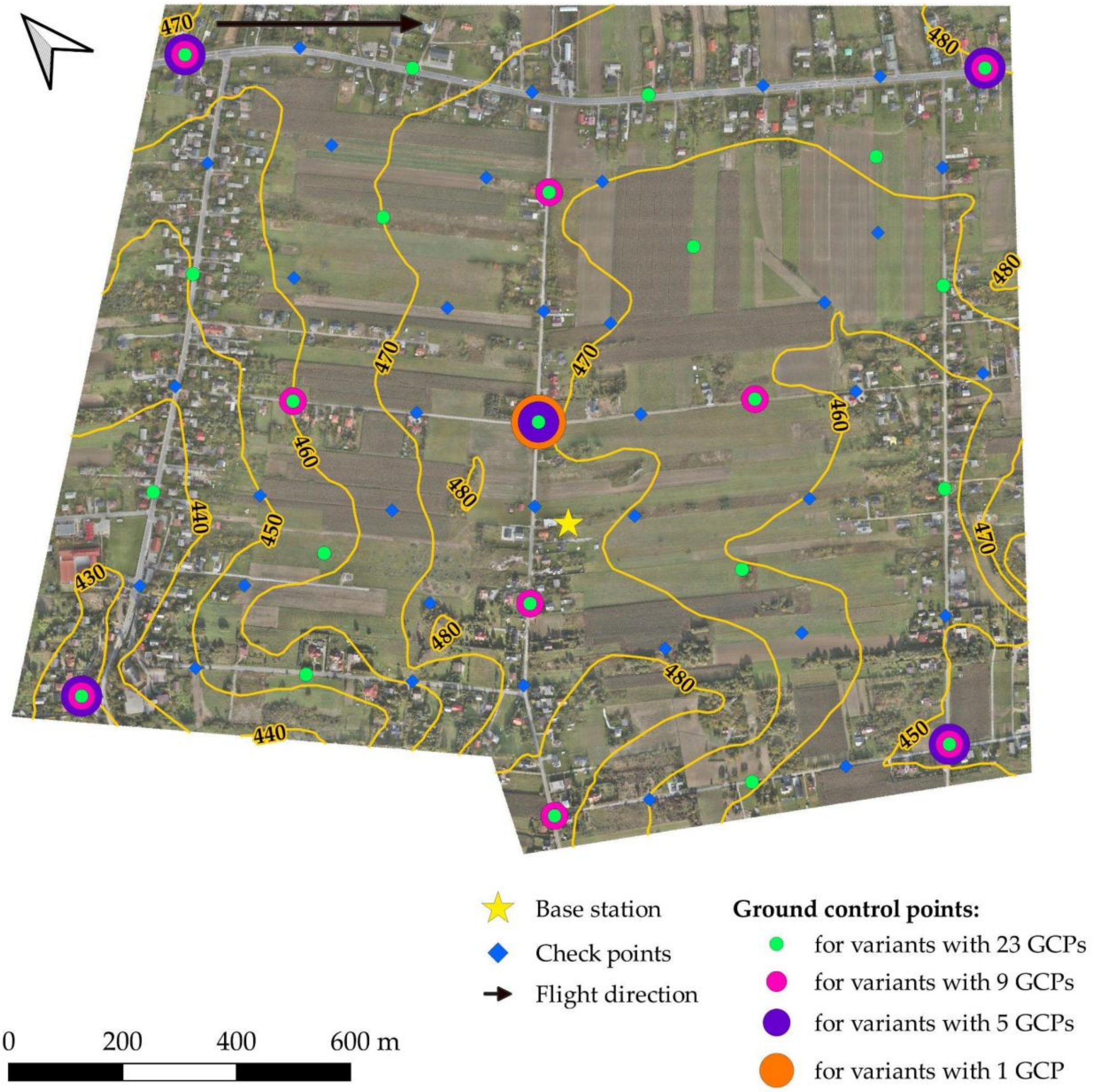

Fig. 1. The UAV-derived orthomosaic showing the land cover and relief of the study area, with marked locations for GCPs (in four tested configurations), CPs, the GNSS base station, and the UAV flight direction.

### 2.2. Tested processing parameters and variants

The research encompassed an analysis of the results from a series of processing variants. All calculations were performed using Agisoft Metashape Professional ver. 2.0.3 (build 16960, formerly PhotoScan). Table 1 summarizes the processing parameters and their values examined for their influence on the accuracy of the results. Every possible combination of the listed parameter values was tested, resulting in a total of 768 variants, which were calculated for data from each measurement series.

Tab. 1. Parameter values used in the processing variants.

| No. | Parameter | Parameter name abbreviation | Number of variants | Tested values |
|---|---|---|---|---|
| 1 | Method of camera projection center coordinates determination | PCMD | 2 | RTN, PPK |
| 2 | Number of GCPs | NGCP | 4 | 1, 5, 9, 23 |
| 3 | GCP coordinate accuracy | GCPA | 2 | 25 mm, 15 mm |
| 4 | Additional corrections | AC | 2 | OFF, ON |
| 5 | Tie point accuracy | TPA | 2 | 1 pixel, 2 pixels |
| 6 | Camera projection center coordinate accuracy | PCA | 3 | 100 mm, 50 mm, estimated |
| 7 | Tie point filtration | TPF | 2 | NO, YES |
| 8 | Transformation | TR | 2 | NO, YES |

For the processing variants that employed the GNSS PPK method to determine the coordinates of camera projection centers, the georeferencing was performed relative to a dedicated physical GNSS base station established on-site (Fig. 1). The coordinates of this station—identical to the one used for the measurements of GCPs and CPs—were constant and equal to the values determined in the static GNSS survey described in Section 2.1. Using this on-site base station ensured short baselines during GNSS PPK processing, with the maximum baseline length not exceeding 1 km. During each measurement series, a Leica GS16 receiver mounted on a tripod at the base station recorded continuous static GNSS observations at a 0.2 s sampling rate. These observations served exclusively for GNSS PPK processing (performed using the REDtoolbox software) and for deriving the camera projection center coordinates within the same reference frame as the GCP and CP coordinates.

In the processing variants, where the coordinates of camera projection centers were determined using the GNSS RTN method, the georeferencing was tied to Virtual Reference Stations (VRS) generated from the ASG-EUPOS network. The nearest ASG-EUPOS stations were located approximately 20 km, 40 km, and 50 km from the study area, as described in Section 2.1. Consequently, the accuracy of the camera projection center coordinates depended on the accuracy with which the VRS positions were determined within the reference frame defined by the GCP and CP coordinates.

Three sets of *a priori* uncertainty estimates for the camera projection center coordinates were considered in the BBA. The first set used accuracy estimations derived directly from the results of the GNSS PPK and GNSS RTN processing, respectively. The other two sets assumed uniform *a priori* uncertainties for all coordinates of every camera projection center, applying values of 50 mm and 100 mm, respectively.

The spatial distribution of GCPs for each processing variant is illustrated in Figure 1. In variants requiring a higher number of GCPs, new points were added without altering the positions of the GCPs used in the variants with fewer points. The *a priori* GCP coordinate accuracy values, listed in Table 1, represent the assumed RMSE for each coordinate. In contrast, marker accuracy—which reflects the precision of manual GCP marking—was kept constant across all processing variants and set to 0.5 px. This value was adopted based on typical manual marking repeatability reported in the literature [43] and was further confirmed through a repeatability test.

In all processing variants, the camera calibration parameters were determined using the Brown model [34]. The following parameters were estimated: principal point position ($c_x$, $c_y$), focal length ($f$), radial distortion parameters ($k_1$, $k_2$, $k_3$, $k_4$), and tangential distortion parameters ($p_1$, $p_2$), with all other parameters set to zero. In variants where the AC were enabled, corrections derived from a 96-parameter model proposed by Tang et al. [37] were also fitted.

The next parameter evaluated in the study was tie point accuracy (TPA). This value reflects the normalized projection accuracy of tie points detected at a scale of 1 px, while tie points detected at other scales exhibit accuracy proportional to their scale. Although TPA = 1 px is commonly used in photogrammetric workflows [43], the exceptionally high spatial resolution of the imagery motivated the inclusion of additional variants with TPA = 2 px to assess its influence on the BBA accuracy.

For a subset of the processing variants, a tie point filtration (TPF) was also conducted following the scheme presented in Figure 2. The objective of this filtration was to remove from the adjustment process TPs that could potentially degrade the accuracy of the BBA.

Finally, for a subset of variants, a seven-parameter similarity transformation (three translations, three rotations, and one scale factor) was applied. This transformation process aimed to remove systematic discrepancies between the reprojected and measured coordinates of the GCPs.

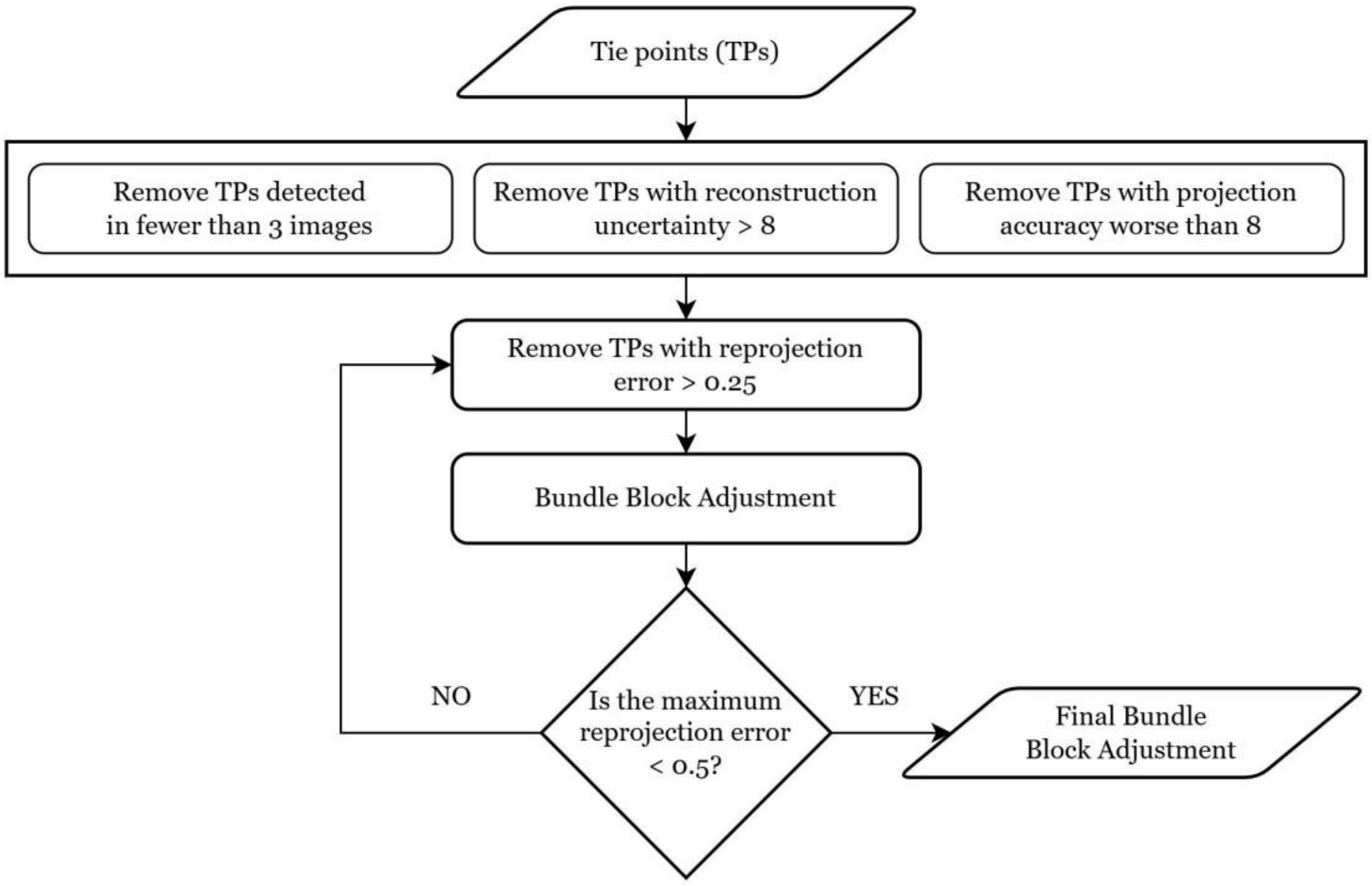

Fig. 2. Workflow of tie points filtration (TPF) using the gradual selection tool in Agisoft Metashape Professional. The reconstruction uncertainty, projection accuracy, reprojection error, and their corresponding thresholds in this tool are unitless [43, 53].

During the processing of data from each measurement series, the BBA was performed at least twice for each processing variant. For the initial image alignment, the key point and tie point limits were set to 40,000 and 10,000 per image, respectively, and the frame camera model was selected. Depending on the specific processing variant, the camera projection center coordinates—determined using either the GNSS RTN or GNSS PPK method—were imported. GCPs, in a number dependent on the processing variant, were marked in the images, and their respective *a priori* accuracies were defined. This procedure resulted in 16 primary processing variants (2 PCMD × 4 NGCP × 2 GCPA), for which an initial BBA was then performed.

During this initial adjustment, all remaining processing parameters were fixed and identical across the variants. The same core set of camera calibration parameters was always estimated: $f$, $c_x$, $c_y$, $k_1$–$k_4$, $p_1$, and $p_2$. The remaining processing parameters were configured as follows: AC = OFF, TPA = 1, PCA = estimated, TPF = NO, Marker Accuracy = 0.5 px, and Camera Accuracy = 10 degrees.

Following the initial BBA, the adjustment results were examined for gross errors in the exterior orientation parameters of individual images. When such errors were detected, the *a priori* values of these parameters for the affected images were excluded from subsequent computations. In each measurement series, this issue affected only a small number of images (<<1% of the total), typically where the exterior orientation parameters were clearly incorrect—often due to an erroneous assignment of the camera projection center from a neighboring image. The analysis involved inspecting the post-adjustment errors in exterior orientation parameters, the number of tie points, and other quality indicators.

Once the results of the initial BBA were reviewed and any gross errors in the exterior orientation parameters removed, a final BBA was performed for each tested processing variant. This second adjustment incorporated the definitive target settings for all parameters specified in the given variant. If a transformation was required for a particular variant, it was applied after the completion of this final BBA.

### 2.3. Evaluation of processing variants and parameter significance

The primary metric for evaluating the performance of the different processing variants was the mean 3D RMSE, calculated from the errors (i.e., differences between the determined and measured coordinates) at the CPs across all analyzed measurement series. The same set of CPs was used for all variants and series (Fig. 1). Based on these criteria, the best-performing processing variant was identified, and an analysis was conducted to determine which parameter values appeared most frequently among the 10% of variants with the lowest 3D RMSE values and the 10% of variants with the highest 3D RMSE values.

The error vectors at the CPs were transformed (by rotating the coordinate system in the horizontal plane) so that their components aligned with the along-track (AL) and across-track (ACR) directions. It enables conducting a processing parameter impact analysis in a coordinate system aligned with the flight path, which is more general than analyzing in the external coordinate system. The mean RMSE values of the AL and ACR components, as well as the elevation (EL) component, were considered as evaluation metrics in those analyses.

Furthermore, the statistical significance of the tested processing parameters was examined using two statistical tests: the t-test and Bartlett's test. The t-test was used to assess whether the values of the tested processing parameters had a significant effect on the mean values of the error components (AL, ACR, and EL) at the CPs, while Bartlett's test was used to evaluate their influence on the variance of these components. Since the t-test only accommodates comparisons between two groups, parameters with more than two possible values were tested using all possible two-value combinations for both tests.

These tests were performed using the results of all processing variants, grouped according to the value of the parameter under evaluation (all other processing parameters had all possible values). A similar analysis was conducted for the subset of the 10% best-performing variants to assess whether the impact of processing parameters on results is similar to their impact on all variants.

### 2.4. Accuracy assessment of displacement, tilt change, and horizontal strain determination

To further analyze the impact of the processing variant on the accuracy of final UAV photogrammetry products (orthomosaics and digital elevation models (DEMs)), ground surface displacements and their derivatives—namely tilt changes and horizontal strains—were also evaluated. The methodology for determining these deformation indicators was adapted from Puniach et al. [44]. This analysis was performed in MATLAB R2022b Update 3 (9.13.0.2126072) (The MathWorks Inc., Natick, Massachusetts, USA) on UAV-derived products generated using two distinct sets of processing parameters.

One of the tested parameter sets (termed the "baseline" variant) was previously used in the study by Puniach et al. [44] and considered sufficiently accurate based on the authors' empirical experience. The baseline variant included the following settings: PCMD = RTN, NGCP = 23, GCPA = 25 mm, AC = ON, TPA = 2 pixels, PCA = 50 mm, TPF = YES, TR = YES.

For comparison, this was contrasted with the parameter set that yielded the lowest mean 3D RMSE across all tested processing variants (termed the "enhanced" variant). This comparison enabled the evaluation of how optimal parameter selection influences the quality of UAV photogrammetry products and allowed for the decomposition of displacement errors into their systematic and random components.

For both tested variants, orthomosaics were generated with a spatial resolution of 10 mm/pixel, and DEMs with a spatial resolution of 20 mm/pixel. Surface deformation indices were calculated for the following time intervals: May 2022 – June 2022, May 2022 – October 2022, October 2022 – January 2023, October 2022 – March 2023, January 2023 – April 2023, January 2023 – July 2023, April 2023 – August 2023, and April 2023 – September 2023. These intervals were selected to represent various combinations of data captured during different vegetation seasons. Data from December 2022 were excluded due to snowfall, which prevented the accurate automatic determination of displacements.

#### 2.4.1. Displacement determination and outlier filtering

Data processing follows the approach proposed by Puniach et al. [44], with its core being the application of image registration algorithms for matching multi-temporal orthomosaics (Fig. 3). The moving registration window (RW) technique was employed to determine displacements. Initially, orthomosaics and DEMs from two measurement series—the base series and the comparison series—were divided into RWs. These RWs were square regions with side lengths equal to 30 m for raster fragments from the base series and were enlarged by 1 m on each side for the comparison series rasters. The centers of adjacent RWs were spaced 30 m from each other, constituting a regular grid. The same data processing approach was applied to each RW.

First, the pixels in RW were classified as either urbanized or non-urbanized. RW classification was performed by segmenting the corresponding orthomosaic fragment from the base series using thresholding of vegetation index computed as (G-B)/(R+G+B), where R, G, B are standardized values (ranging from 0 to 1) for the red, green, and blue channels. The threshold value of 0.019 was adopted from Puniach et al. [54]. The product of such processing was an RW of weight raster, with urbanized pixels assigned a weight of 1 and non-urbanized pixels assigned a weight of 0. If the percentage of pixels within the analyzed RW classified as urbanized met or exceeded 15%, the RW was classified as urbanized. For RWs classified as non-urbanized, displacements were not calculated due to their unreliability.

The Weighted Normalized Cross Correlation (WNCC) algorithm [55] was used to determine horizontal displacement components ($d_n$, $d_e$). This method applies weights to indicate which orthomosaic pixels are relevant for matching.

The next step involved determining vertical displacements ($d_h$) using the DEMs from the base and comparison series. First, the RW of DEM from the comparison series was translated based on the previously determined horizontal displacements. Next, the heights of corresponding DEM pixels within the RWs from both series were subtracted. To reduce the influence of vegetation, height differences for pixels assigned a weight of 0 were excluded from the calculation. If the median absolute deviation of the calculated height differences within the RW was less than or equal to 35 mm, the vertical component of the displacement for that RW was set to the median of these differences. If not, the vertical displacement for that RW was not recorded.

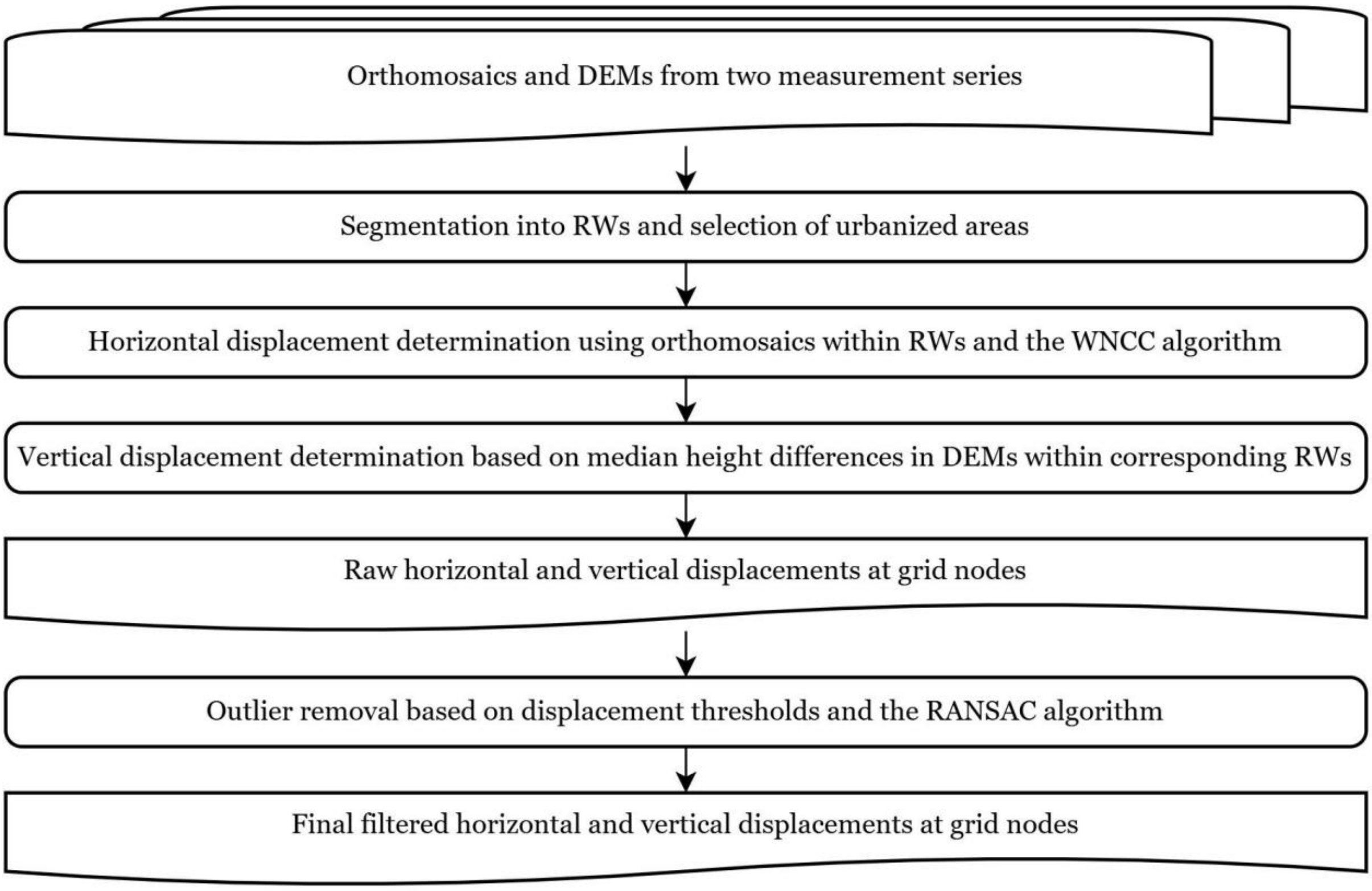

Fig. 3. Flowchart for determining displacements.

The displacement values (both horizontal and vertical) determined constituted the raw displacements, which may contain outliers. Therefore, a method for removing them (filtration) was applied following the approach proposed by Puniach et al. [44].

Removing outliers involved three steps and was performed separately for horizontal and vertical displacements. The initial step of removing outliers was to eliminate horizontal or vertical displacement vectors whose lengths exceeded the maximum expected horizontal displacement (0.6 m) and the maximum expected vertical displacement (2.0 m).

The second filtration step used the Random Sample Consensus (RANSAC) method [56] combined with first-degree polynomial surface fitting to account for the local shape of the displacement surface. This RANSAC-based filtration involved dividing the study area into non-overlapping fragments called filtration windows (FWs) of size 50 m x 50 m, and the polynomial was fitted into a 150 m x 150 m patch for robustness. Outlier detection was performed independently for each displacement component ($d_n$, $d_e$, $d_h$) within each FW. For horizontal displacements, both components were considered outliers if at least one component was identified as such. To identify inliers, the distance between the fitted polynomial and each data point (i.e., displacement component) within the FW was compared to the limit (3 cm for horizontal displacements and 5 cm for vertical displacements).

The final filtration step involved finding and removing isolated displacement vectors. For this purpose, a radius for searching neighboring nodes with determined displacements was defined as 100 m. The final product of this processing stage consists of filtered horizontal and vertical displacements.

#### 2.4.2. Tilt change and horizontal strain determination

The tilt changes and horizontal strains used, respectively, filtered vertical and horizontal displacements as input. Given that the displacements were distributed on a regular grid parallel to the coordinate system axes, the components of tilt changes and horizontal strains were determined along these axes. These values were calculated as the difference in displacements determined at the nearest grid points in the directions of N and E axes, divided by the distance between nodes. They can be attributed to points located halfway between the grid nodes:

$$t_{\mathrm{n}}\left(n+\frac{\mathrm{S}}{2},e\right)=\frac{d_{\mathrm{h}}(n+\mathrm{S},e)-d_{\mathrm{h}}(n,e)}{\mathrm{S}},$$

$$t_{\mathrm{e}}\left(n,e+\frac{\mathrm{S}}{2}\right)=\frac{d_{\mathrm{h}}(n,e+\mathrm{S})-d_{\mathrm{h}}(n,e)}{\mathrm{S}},$$

$$\varepsilon_{\mathrm{n}}\left(n+\frac{\mathrm{S}}{2},e\right)=\frac{\sqrt{(d_{\mathrm{n}}(n+\mathrm{S},e)-d_{\mathrm{n}}(n,e)+\mathrm{S})^2+\left(d_{\mathrm{e}}(n+\mathrm{S},e)-d_{\mathrm{e}}(n,e)\right)^2}-\mathrm{S}}{\mathrm{S}},$$

$$\varepsilon_{\mathrm{e}}\left(n,e+\frac{\mathrm{S}}{2}\right)=\frac{\sqrt{(d_{\mathrm{e}}(n,e+\mathrm{S})-d_{\mathrm{e}}(n,e)+\mathrm{S})^2+\left(d_{\mathrm{n}}(n,e+\mathrm{S})-d_{\mathrm{n}}(n,e)\right)^2}-\mathrm{S}}{\mathrm{S}},$$

where $n$ and $e$ are grid node coordinates, $t_{\mathrm{n}}$ and $t_{\mathrm{e}}$ are tilt change components, $\varepsilon_{\mathrm{n}}$ and $\varepsilon_{\mathrm{e}}$ are horizontal strain components, and S is the distance between adjacent RWs.

#### 2.4.3. Error assessment of deformation indices

The basic statistics used to evaluate the deformation indices included the total number of calculated values for each index, RMSE, and the percentage of outliers. Thresholds for classifying values as outliers were defined as follows: 9 cm for horizontal displacements, 15 cm for vertical displacements, 5 mm/m for tilt change components, and 3 mm/m for horizontal strain components. These outliers were excluded from the RMSE calculation, as even a small number of extreme values could significantly distort the results and hinder meaningful interpretation. Such errors were not treated in the same manner as random or systematic errors, as their distribution characteristics are fundamentally different.

To further distinguish between systematic and random error components in the displacement data, a methodology was applied using thin plate smoothing splines [57] and cross-validation. A smoothing parameter of 1 was used, corresponding to a cubic spline interpolant. A 10-fold cross-validation was performed by randomly dividing the dataset into ten equal subsets. For each iteration, 9/10 of the data were used to interpolate values for the remaining 1/10. The interpolated values represented the systematic error component—i.e., the portion of the error spatially correlated with neighboring observations. The random error component was defined as the difference between the total error and its systematic portion at each node.

Accuracy assessment and error analysis were conducted independently for each displacement component, across all data series, and for both the enhanced and baseline variants. To summarize the systematic error component, the Median Absolute Error (MAE) was calculated across all time intervals, with results reported separately for each processing variant. MAE was selected to reduce the influence of extreme values, such as those occasionally observed in non-urban areas misclassified as urban. For the random error component, RMSE values were computed based on results from all time intervals, also reported separately for each processing variant.

## 3. Results

This section presents the results of analyses assessing the influence of individual processing parameters on adjustment accuracy and identifying the best-performing processing variant among those tested. It also includes an evaluation of the accuracy of ground surface deformation indices derived from UAV-based rasters generated using different parameter configurations.

### 3.1. Assessment of processing variants and parameter significance

After calculating results for all processing variants for every time interval, the errors on the CPs were analyzed. For this purpose, the processing variants were sorted according to their 3D RMSE on the CPs (Appendix 1). It was first noted that the parameter settings generally have a much greater impact on the elevation errors (EL) than on the horizontal components (AL and ACR) (Appendix 1). As a result, ranking the processing variants by 3D RMSE is nearly identical to a ranking based solely on the elevation RMSE (RMSE EL).

Figure 4 presents the proportion of parameter values within the worst 10% of variants exhibiting the highest 3D RMSE on the CPs. In these underperforming variants, the mean 3D RMSE values, calculated from all analyzed intervals, ranged from 64.8 mm to 303.4 mm. The analysis of these worst processing variants confirms almost unequivocally that using only one GCP and not applying the additional corrections—particularly in combination—leads to poor results. This finding is further supported by the density plots of errors for the CPs (based on kernel density estimation), which illustrate the effect of the number of GCPs (comparing one vs. five) and the application of additional corrections (Fig. 5).

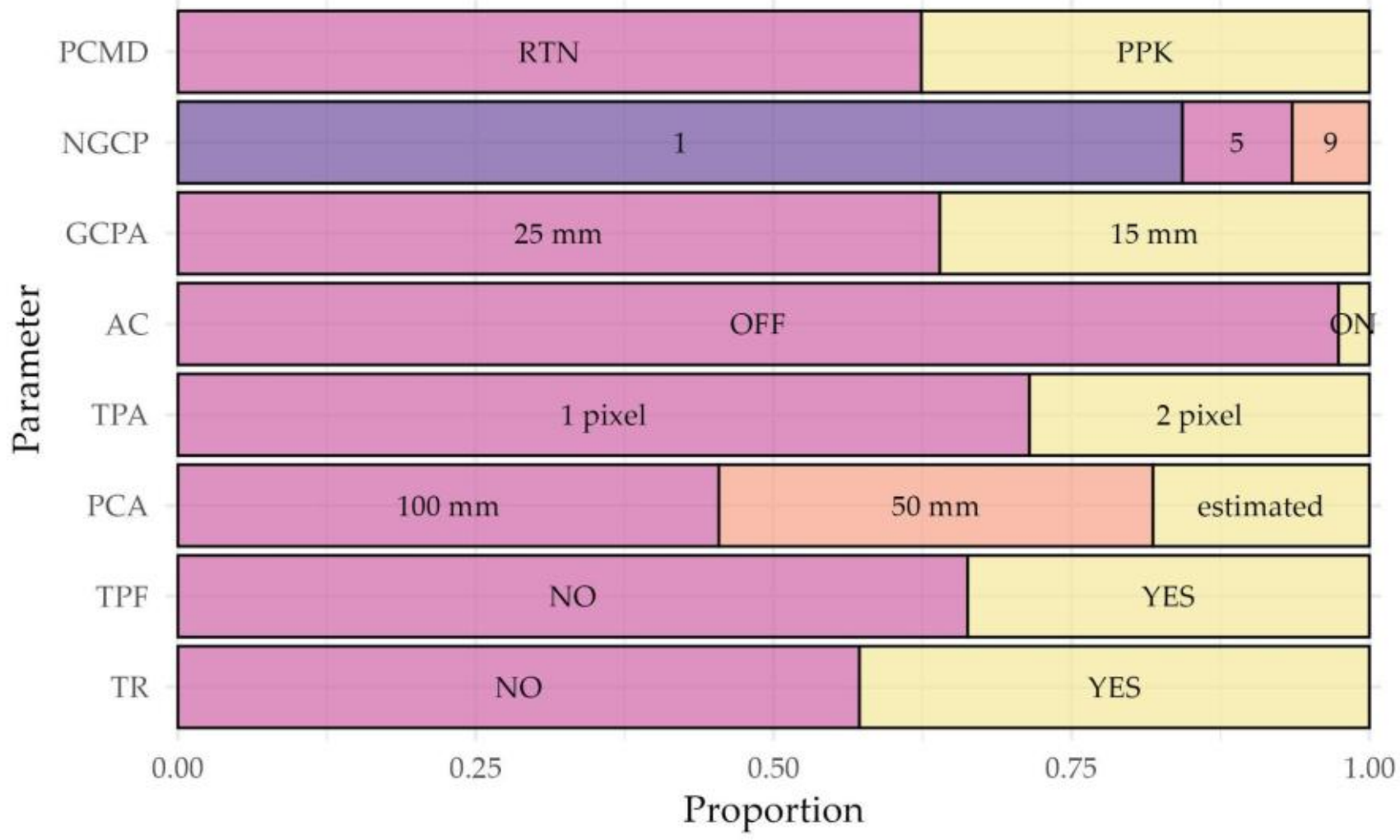

Fig. 4. Proportion of parameter values among the worst 10% (highest 3D RMSE) of processing variants based on the mean 3D RMSE calculated for the CPs in all measurement series

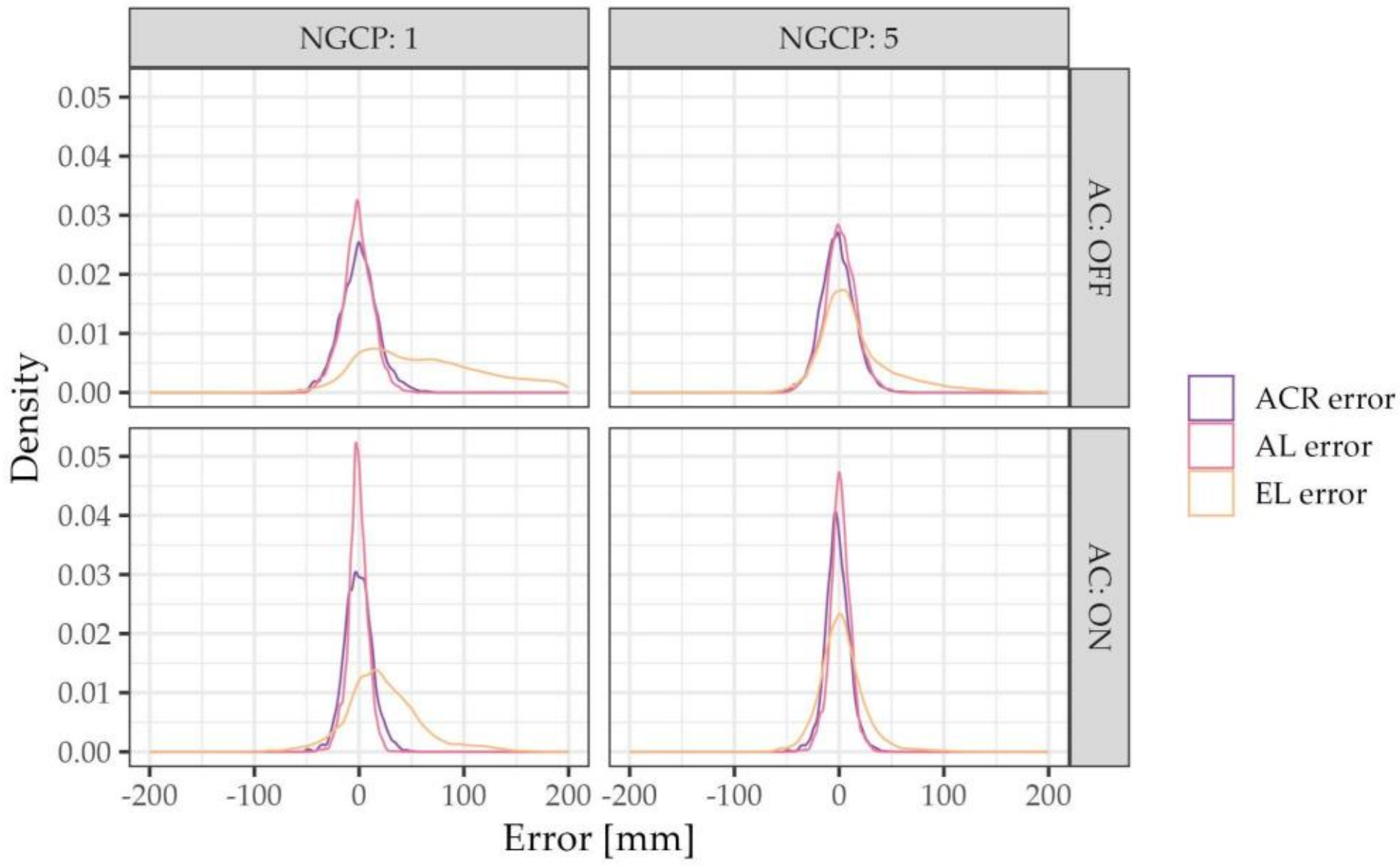

Fig. 5. Density plots of errors for the CPs across all processing variants and measurement series, illustrating the impact of increasing NGCP from 1 to 5 and the effect of enabling AC

On the other hand, the mean 3D RMSE for the top 10% of variants ranged from approximately 15.9 mm to 19.8 mm. The proportion of parameter value occurrence for these best-performing variants is presented in Figure 6. An analysis of this group reveals key findings:

- In all of the best-performing variants, the additional corrections were used, and the number of GCPs was always greater than one.
- In over 90% of these top variants, GNSS PPK was the method used for determining the camera projection center coordinates.

Further analysis indicates that applying a transformation is beneficial, as it was used in over 70% of the top processing variants. Similarly, a tie point accuracy of 2 pixels appears to be preferable, as this value was used in over 70% of the top processing variants.

Conversely, the analysis of these high-performing variants suggests that the specific value assumed for GCP coordinate accuracy (15 mm or 25 mm) has no significant impact on the outcome. Likewise, the application of tie-point filtration was not a determining factor in achieving the best results.

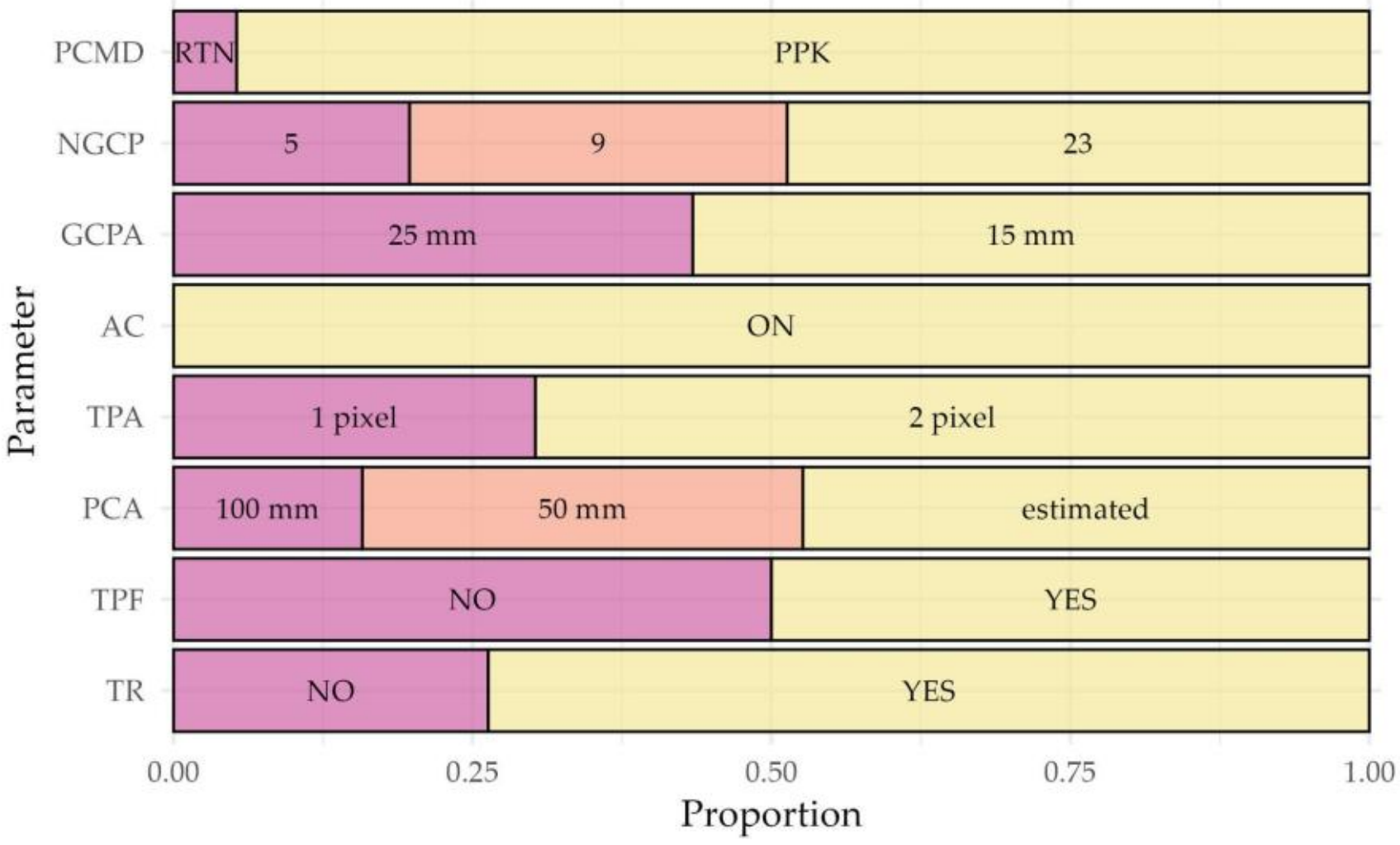

Fig. 6. Proportion of parameter values among the best 10% (lowest 3D RMSE) of processing variants based on the mean 3D RMSE calculated for the CPs in all measurement series

The significance of the method for determining camera projection center coordinates and the application of a transformation for the top 10% of variants is well-illustrated by the histograms of errors calculated for the CPs, broken down by these parameter values (Fig. 7). However, it must be emphasized again that these parameters only have a significant impact when the foundational conditions of using more than one GCP and applying the additional corrections are met.

Analyzing the impact of parameter settings on the components of CP errors reveals that the application of additional corrections is significant for all components (Fig. 8). To minimize errors in the AL component, it is crucial to use GNSS PPK for determining camera projection centers and to apply the *a priori* weights for these observations as estimated from the PPK calculations (Fig. 8 (a)). For the ACR component, the aforementioned parameters are less critical, but the application of a transformation proves to be very important (Fig. 8 (b)). For the EL component, the key factor is the use of GNSS PPK for determining camera projection centers, with a high number of GCPs also being very beneficial (Fig. 8 (c)). Therefore, each error component is critically influenced by a slightly different set of parameters. Fortunately, the optimal settings for minimizing errors in each component are not mutually exclusive.

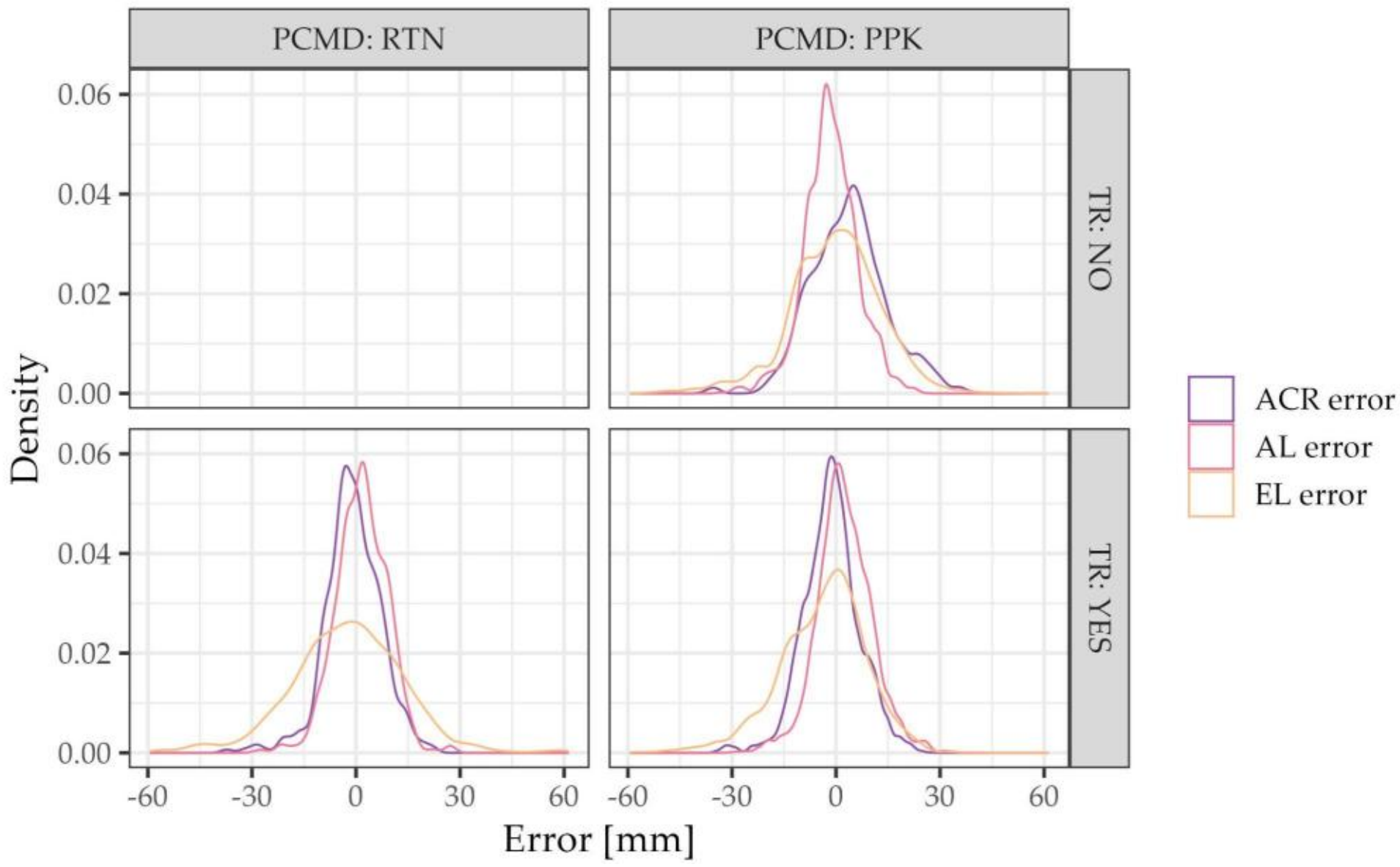

Fig. 7. Density plots of errors for the CPs across 10% of the top processing variants and all measurement series, illustrating the impact of the method for determining camera projection center coordinates and the application of a transformation. In the top 10% variants, there was no case where, simultaneously, the camera projection center coordinates were determined using GNSS RTN and transformation had not been applied.

Statistical analysis, encompassing t-tests for mean values and Bartlett's tests for variances across all processing variants, leads to a general conclusion: at the 5% significance level ($\alpha$=0.05), nearly each tested parameter demonstrated a statistically significant effect on both the mean and the variance of the CP errors (Tab. 2 and 3). An exception was only observed for the ACR error component. For this component, the t-test did not indicate significant differences in the mean errors when comparing results obtained using 9 and 23 GCPs. Similarly, the *a priori* accuracy of GCP coordinates (GCPA) did not have a statistically significant impact on the mean values of the ACR error component. Furthermore, an analysis of error variances for the ACR component revealed that the number of GCPs (provided that more than one GCP is used, i.e., 5, 9, or 23) does not significantly affect the variability of the error for this component.

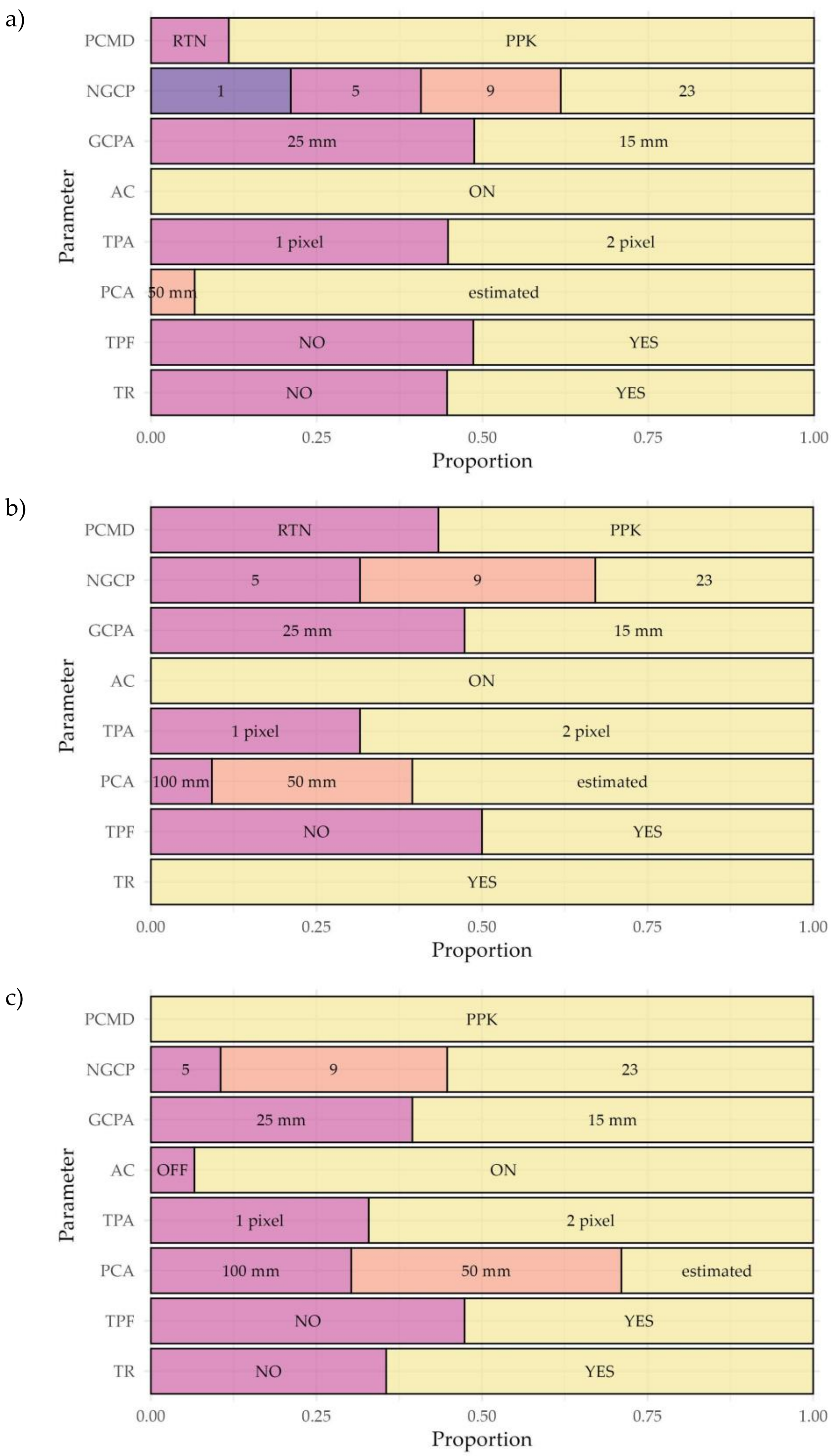

Fig. 8. Proportion of parameter values among the best 10% of processing variants based on: a) the mean AL RMSE, b) the mean ACR RMSE, c) the mean EL RMSE calculated for the CPs in all measurement series

Tab. 2. Comparison of mean errors (in AL, ACR, and EL components) at CPs between all processing variants with different values of selected processing parameters. Reported *p*-values are based on independent two-sample t-tests.

| Parameter | | | AL errors | | | ACR errors | | | EL errors | | |
|---|---|---|---|---|---|---|---|---|---|---|---|
| **Name** | **Value #1** | **Value #2** | **Mean #1 [mm]** | **Mean #2 [mm]** | ***p*-value** | **Mean #1 [mm]** | **Mean #2 [mm]** | ***p*-value** | **Mean #1 [mm]** | **Mean #2 [mm]** | ***p*-value** |
| PCMD | PPK | RTN | 0.1 | 0.7 | 0.000 | 2.3 | -4.1 | 0.000 | 14.9 | 21.2 | 0.000 |
| NGCP | 1 | 5 | -1.6 | 0.9 | 0.000 | -0.4 | -1.6 | 0.000 | 54.9 | 9.7 | 0.000 |
| NGCP | 1 | 9 | -1.6 | 1.8 | 0.000 | -0.4 | -0.8 | 0.000 | 54.9 | 6.0 | 0.000 |
| NGCP | 1 | 23 | -1.6 | 0.4 | 0.000 | -0.4 | -0.8 | 0.000 | 54.9 | 1.7 | 0.000 |
| NGCP | 5 | 9 | 0.9 | 1.8 | 0.000 | -1.6 | -0.8 | 0.000 | 9.7 | 6.0 | 0.000 |
| NGCP | 5 | 23 | 0.9 | 0.4 | 0.000 | -1.6 | -0.8 | 0.000 | 9.7 | 1.7 | 0.000 |
| NGCP | 9 | 23 | 1.8 | 0.4 | 0.000 | -0.8 | -0.8 | **0.605** | 6.0 | 1.7 | 0.000 |
| GCPA | 15 mm | 25 mm | 0.4 | 0.3 | 0.006 | -0.9 | -0.9 | **0.140** | 14.2 | 21.9 | 0.000 |
| AC | OFF | ON | 0.5 | 0.3 | 0.000 | -0.7 | -1.1 | 0.000 | 29.6 | 6.6 | 0.000 |
| TPA | 1 pixel | 2 pixels | 0.3 | 0.4 | 0.006 | -0.6 | -1.2 | 0.000 | 25.7 | 10.5 | 0.000 |
| PCA | 50 mm | 100 mm | 0.4 | 0.3 | 0.045 | -1.0 | -0.4 | 0.000 | 18.2 | 23.3 | 0.000 |
| PCA | 50 mm | estimated | 0.4 | 0.5 | 0.028 | -1.0 | -1.3 | 0.000 | 18.2 | 12.8 | 0.000 |
| PCA | 100 mm | estimated | 0.3 | 0.5 | 0.000 | -0.4 | -1.3 | 0.000 | 23.3 | 12.8 | 0.000 |
| TPF | NO | YES | 0.3 | 0.4 | 0.015 | -0.7 | -1.1 | 0.000 | 23.7 | 12.5 | 0.000 |
| TR | NO | YES | -1.3 | 2.0 | 0.000 | -0.6 | -1.2 | 0.000 | 24.6 | 11.5 | 0.000 |

Tab. 3. Comparison of the standard deviation (SD) of errors (in AL, ACR, and EL components) at CPs between all processing variants with different values of selected processing parameters. Reported *p*-values are based on Bartlett's tests for differences in variances.

| Parameter | | | AL errors | | | ACR errors | | | EL errors | | |
|---|---|---|---|---|---|---|---|---|---|---|---|
| **Name** | **Value #1** | **Value #2** | **SD #1 [mm]** | **SD #2 [mm]** | ***p*-value** | **SD #1 [mm]** | **SD #2 [mm]** | ***p*-value** | **SD #1 [mm]** | **SD #2 [mm]** | ***p*-value** |
| PCMD | PPK | RTN | 12.4 | 12.6 | 0.000 | 14.0 | 14.3 | 0.000 | 44.8 | 53.6 | 0.000 |
| NGCP | 1 | 5 | 12.3 | 12.9 | 0.000 | 15.9 | 14.0 | 0.000 | 76.8 | 32.3 | 0.000 |
| NGCP | 1 | 9 | 12.3 | 12.6 | 0.000 | 15.9 | 13.9 | 0.000 | 76.8 | 25.7 | 0.000 |
| NGCP | 1 | 23 | 12.3 | 11.9 | 0.000 | 15.9 | 14.0 | 0.000 | 76.8 | 18.9 | 0.000 |
| NGCP | 5 | 9 | 12.9 | 12.6 | 0.000 | 14.0 | 13.9 | **0.328** | 32.3 | 25.7 | 0.000 |
| NGCP | 5 | 23 | 12.9 | 11.9 | 0.000 | 14.0 | 14.0 | **0.871** | 32.3 | 18.9 | 0.000 |
| NGCP | 9 | 23 | 12.6 | 11.9 | 0.000 | 13.9 | 14.0 | **0.254** | 25.7 | 18.9 | 0.000 |
| GCPA | 15 mm | 25 mm | 12.5 | 12.6 | 0.005 | 14.4 | 14.5 | 0.006 | 41.7 | 55.9 | 0.000 |
| AC | OFF | ON | 15.0 | 9.4 | 0.000 | 16.7 | 11.9 | 0.000 | 62.8 | 26.4 | 0.000 |
| TPA | 1 pixel | 2 pixels | 13.2 | 11.8 | 0.000 | 15.1 | 13.8 | 0.000 | 60.5 | 34.2 | 0.000 |
| PCA | 50 mm | 100 mm | 12.9 | 14.2 | 0.000 | 14.5 | 15.9 | 0.000 | 46.7 | 60.5 | 0.000 |
| PCA | 50 mm | estimated | 12.9 | 10.0 | 0.000 | 14.5 | 12.7 | 0.000 | 46.7 | 38.1 | 0.000 |
| PCA | 100 mm | estimated | 14.2 | 10.0 | 0.000 | 15.9 | 12.7 | 0.000 | 60.5 | 38.1 | 0.000 |
| TPF | NO | YES | 12.8 | 12.2 | 0.000 | 14.7 | 14.2 | 0.000 | 57.8 | 38.8 | 0.000 |
| TR | NO | YES | 12.1 | 12.6 | 0.000 | 15.6 | 13.2 | 0.000 | 50.1 | 48.0 | 0.000 |

Statistical significance does not necessarily imply practical significance, which is less formally defined and often more subjective. In the following analysis, the assumption was taken that a practically significant difference is defined as a case where the difference in mean error or mean standard deviations of errors calculated for two variant groups (differing by a single parameter setting) exceeds 2 mm. From this perspective, when analyzing results across all processing variants, it is evident that all tested parameters influence both the mean value and the standard deviation of the EL component of the CP errors (Tab. 2 and 3). The most impactful parameter in reducing both metrics is the use of more than one GCP (i.e., 5, 9, or 23 instead of 1). This setting leads to at least a fivefold reduction in the mean error and more than a twofold reduction in the standard deviation. It also ensures that relatively good results are obtained regardless of the values of the other parameters. The next most influential parameters in reducing both the mean and variability of the EL error component are the use of additional corrections and the accuracy setting for tie points.

The influence of parameter settings on the AL and ACR components of the CP errors is noticeably smaller and primarily reflected in differences in standard deviation rather than in the mean error. Considering the assumption on practical significance, the following factors can be considered practically important: enabling additional corrections, utilizing estimated values of the camera projection center accuracy, and applying transformation. The importance of using more than one GCP is on the edge of being practically important (1.9-2.0 mm differences in mean standard deviation).

In the case of the mean values of error components, the key aspect is not just the reduction of these metrics, but how closely the mean error approaches zero. A non-zero mean error indicates a systematic bias in the results relative to the ground truth data. It is worth noting that for the AL and ACR components, this issue is virtually negligible (Table 2), with all parameter settings yielding a mean error close to 0 mm (±4 mm). However, the situation differs for the EL component. Among all tested parameters, only the use of 23 GCPs—regardless of other parameter settings—brings the mean EL error component close to zero, with a value of 1.7 mm in this case.

For the top 10% of processing variants, only the mean standard deviations and results of Bartlett's test for differences in variance were presented (Tab. 4), as the mean errors for all components at the CPs were close to 0 mm (±4 mm), which is a desirable effect. Even within this limited subset of variants, Bartlett's test results indicate that each parameter is statistically significant for at least one error component. However, it is worth noting that for the EL component, the number of GCPs does not significantly affect the results as long as more than one is used (i.e., 5, 9, or 23), nor does the *a priori* accuracy of GCPs or tie point accuracy. When considering practical significance—defined here in terms of noticeable differences in the mean standard deviation of EL error components—the parameter with the strongest influence is the method used to determine camera projection center coordinates (PPK vs. RTN). The use of GNSS PPK results in a reduction of the mean standard deviation of EL components of errors by 3.3 mm. For the AL and ACR error components, the most practically relevant factors appear to be the *a priori* accuracy of the camera projection center coordinates (with estimated values improving accuracy) and the application of transformation. The conclusions from the analysis resulting from Bartlett's tests for the top 10% of processing variants are consistent with the conclusions from the analysis of the parameter values for these variants (Fig. 8).

Tab. 4. Comparison of the standard deviation (SD) of errors (in AL, ACR, and EL components) at CPs between the top 10% of processing variants with different values of selected processing parameters. Reported $p$-values are based on Bartlett's tests for differences in variances.

| **Parameter** | | | **AL errors** | | | **ACR errors** | | | **EL errors** | | |
|---|---|---|---|---|---|---|---|---|---|---|---|
| **Name** | **Value #1** | **Value #2** | **SD #1 [mm]** | **SD #2 [mm]** | ***p*-value** | **SD #1 [mm]** | **SD #2 [mm]** | ***p*-value** | **SD #1 [mm]** | **SD #2 [mm]** | ***p*-value** |
| PCMD | PPK | RTN | 8.2 | 7.4 | 0.000 | 9.4 | 8.2 | 0.000 | 12.8 | 16.1 | 0.000 |
| NGCP | 5 | 9 | 8.2 | 8.3 | 0.002 | 8.5 | 9.1 | 0.000 | 13.1 | 13.0 | **0.465** |
| NGCP | 5 | 23 | 8.2 | 7.9 | **0.251** | 8.5 | 9.7 | 0.000 | 13.1 | 13.0 | **0.629** |
| NGCP | 9 | 23 | 8.3 | 7.9 | 0.000 | 9.1 | 9.7 | 0.000 | 13.0 | 13.0 | **0.724** |
| GCPA | 15 mm | 25 mm | 8.2 | 8.1 | **0.101** | 9.5 | 9.0 | 0.000 | 13.0 | 13.1 | **0.410** |
| TPA | 1 pixel | 2 pixels | 8.1 | 8.1 | **0.264** | 8.7 | 9.5 | 0.000 | 13.1 | 13.0 | **0.502** |
| PCA | 50 mm | 100 mm | 8.6 | 9.7 | 0.000 | 9.8 | 10.2 | 0.000 | 12.7 | 12.3 | 0.023 |
| PCA | 50 mm | estimated | 8.6 | 7.1 | 0.000 | 9.7 | 8.6 | 0.000 | 12.7 | 13.5 | 0.000 |
| PCA | 100 mm | estimated | 9.7 | 7.1 | 0.000 | 10.2 | 8.6 | 0.000 | 12.3 | 13.5 | 0.000 |
| TPF | NO | YES | 8.1 | 8.2 | **0.231** | 9.2 | 9.5 | 0.000 | 13.2 | 12.8 | 0.001 |
| TR | NO | YES | 7.8 | 8.0 | 0.002 | 11.0 | 8.2 | 0.000 | 13.2 | 12.9 | 0.019 |

* some parameter values—specifically, NGCP = 1 and AC = OFF—were not present among the top 10% of processing variants and are therefore not included in the table

### 3.2. Accuracy assessment of deformation indices

Taking into account the adopted criterion for selecting the best processing variant (minimum 3D RMSE), the optimal configuration included the following parameter settings: PCMD = PPK, NGCP = 23, GCPA = 15 mm, AC = ON, TPA = 2 pixels, PCA = estimated, TPF = NO, TR = YES. In the following subsection, this variant is referred to as an enhanced variant.

The differences between the baseline (subsection 2.4) and enhanced variants involved the following parameters, listed in order of their impact on the results:

- Method of camera projection center coordinates determination (PCMD): baseline: RTN, enhanced: PPK,
- Camera projection center coordinate accuracy (PCA): baseline: 50 mm, enhanced: estimated,
- GCP coordinate accuracy (GCPA): baseline: 25 mm, enhanced: 15 mm,
- Tie point filtration (TPF): baseline: ON, enhanced: OFF.

It is worth emphasizing that in previous studies focused on displacement determination accuracy [44], the parameters (baseline variant) did not differ substantially from the variant identified through in-depth analysis as the best-performing (enhanced). For several key parameters, both baseline and enhanced variants utilized settings that enabled relatively high accuracy results (NGCP = 23, AC = ON, TR = YES). Therefore, the results presented here help illustrate the extent to which the accuracy of deformation indices determination can still be improved by optimizing the settings of other parameters.

Table 5 summarizes the basic statistics for the displacements determined from UAV photogrammetry products generated using these two processing variants. While the changes in parameters had a negligible effect on the number of determined displacements and the percentage of outliers, their influence on the RMSE of the determined displacement values may have some practical significance, particularly for the vertical components. Figure 9 presents histograms of the errors for the calculated displacement components. These clearly show that the error distributions for the enhanced variant are more concentrated around zero, an effect that is especially pronounced for the vertical displacements.

Tab. 5. Summary of the accuracy assessment of displacements determined for the JERZ area, based on rasters generated with two different sets of processing parameters. The table provides the RMSE, the number of analyzed displacement components, and the percentage of outliers retained after data filtration, determined cumulatively for all analyzed periods.

| Variant | Number of displacement components | | Outlier percentage [%] | | RMSE [mm] | | |
|---|---|---|---|---|---|---|---|
| | $d_n/d_e$ | $d_h$ | $d_n/d_e$ | $d_h$ | $d_n$ | $d_e$ | $d_h$ |
| Baseline | 7798 | 7271 | 0.8 | 0.0 | 10 | 8 | 21 |
| Enhanced | 7825 | 7273 | 0.7 | 0.0 | 7 | 7 | 14 |

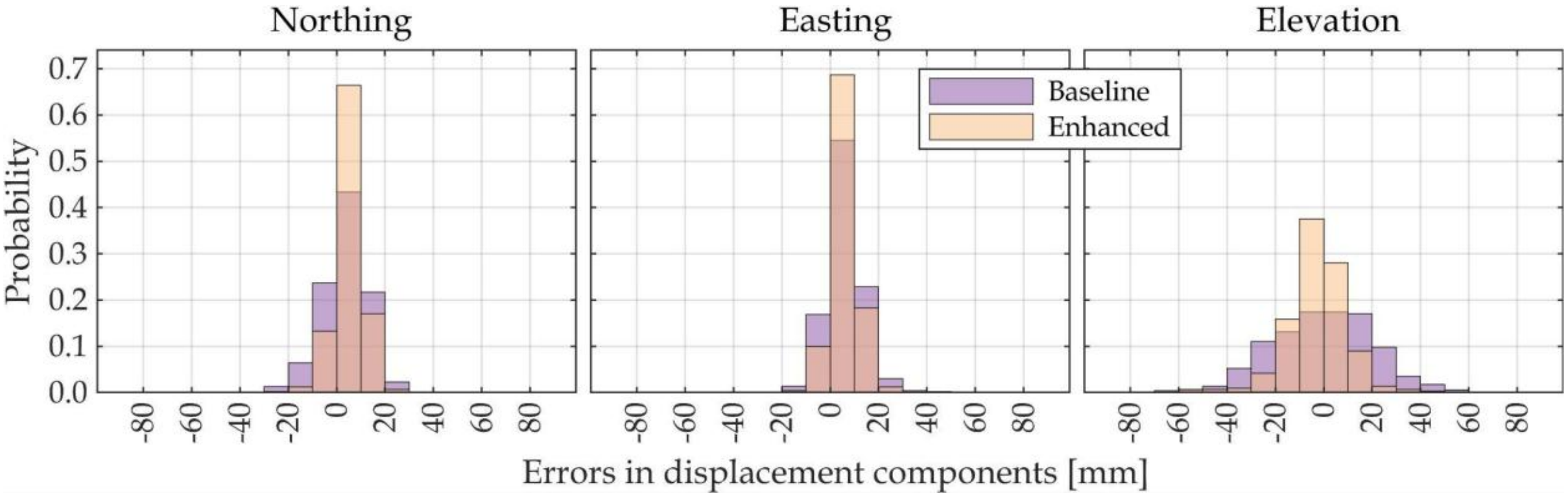

Fig. 9. Histograms of the displacement component errors determined for the JERZ area, based on rasters generated with two different sets of processing parameters. The figure shows the distribution of the displacement component errors determined cumulatively for all analyzed periods.

To gain further insight into the reasons for the reduced RMSE, the errors were analyzed into their systematic and random components. In this context, the systematic component is defined as the error values that are spatially correlated with values at neighboring nodes. As such, these systematic errors do not have a constant value for a given time interval but exhibit gradual spatial variation. An example visualization of these systematic error components, obtained from processing UAV photogrammetry data generated for both the baseline and enhanced variants within a single time interval, is presented in Figure 10. It shows the slow change of the systematic error component with position, with its magnitude, on average, being larger for the baseline variant than for the enhanced variant.

The histograms of the random error components (Fig. 11) are nearly identical for both variants. In contrast, the histograms for the systematic error components (Fig. 12) show a clear difference: the distributions for the enhanced variant are more concentrated around zero than those for the baseline variant. This effect is most prominent for the vertical component of the determined displacements, mirroring the RMSE results for total errors.

These visual observations are confirmed by the summary statistics. For both variants, the random error for each displacement component is at a consistent level of 6–7 mm (RMSE). However, the MAE of the systematic error components is significantly smaller for the enhanced variant. For the horizontal components, the MAE of systematic errors in the enhanced variant is approximately 2 mm for each axis, compared to 4 mm and 6 mm in the baseline variant. For the vertical component, the MAE of the systematic errors is 7 mm for the enhanced variant, whereas for the baseline variant it is 14 mm—a twofold increase.

These observations unequivocally indicate that the proper selection of processing parameters for very high-resolution UAV photogrammetry data (i.e., 10 mm/pixel) enables a significant reduction in the systematic error components. The source of these systematic errors should be sought within the UAV photogrammetry products used to derive the displacements, rather than in the displacement determination method itself, which has an accuracy of 1 pixel for each horizontal component.

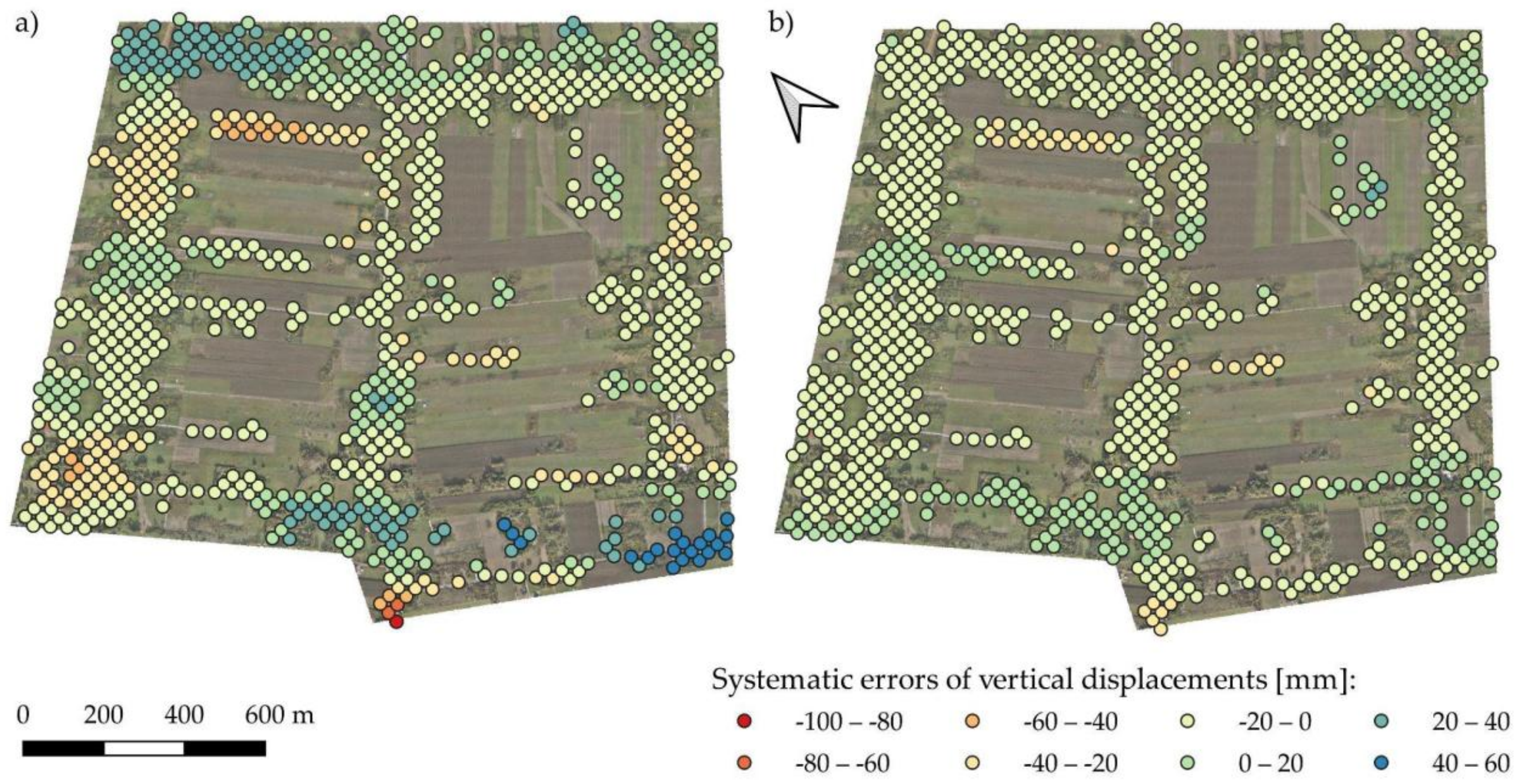

Fig. 10. Systematic errors in vertical displacements for the JERZ area between October 2022 and January 2023, based on rasters generated using two different variants: (a) baseline, (b) enhanced.

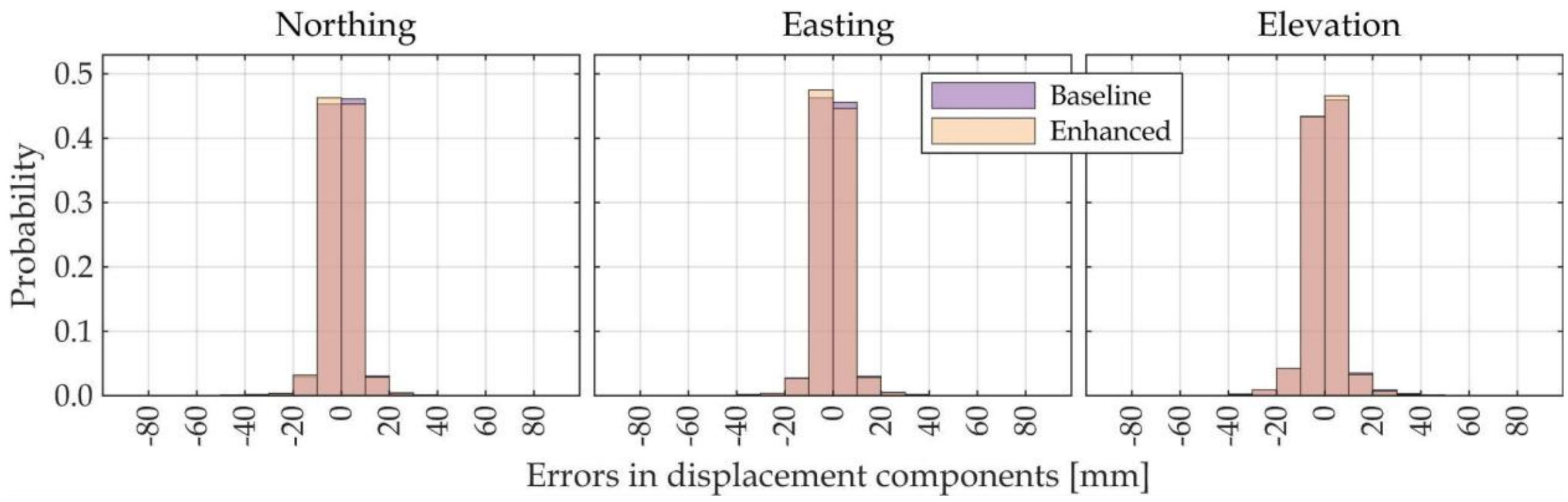

Fig. 11. Histograms of the **random** displacement component errors determined for the JERZ area, based on rasters generated with two different sets of processing parameters. The figure shows the distribution of the displacement component errors determined cumulatively for all analyzed periods.

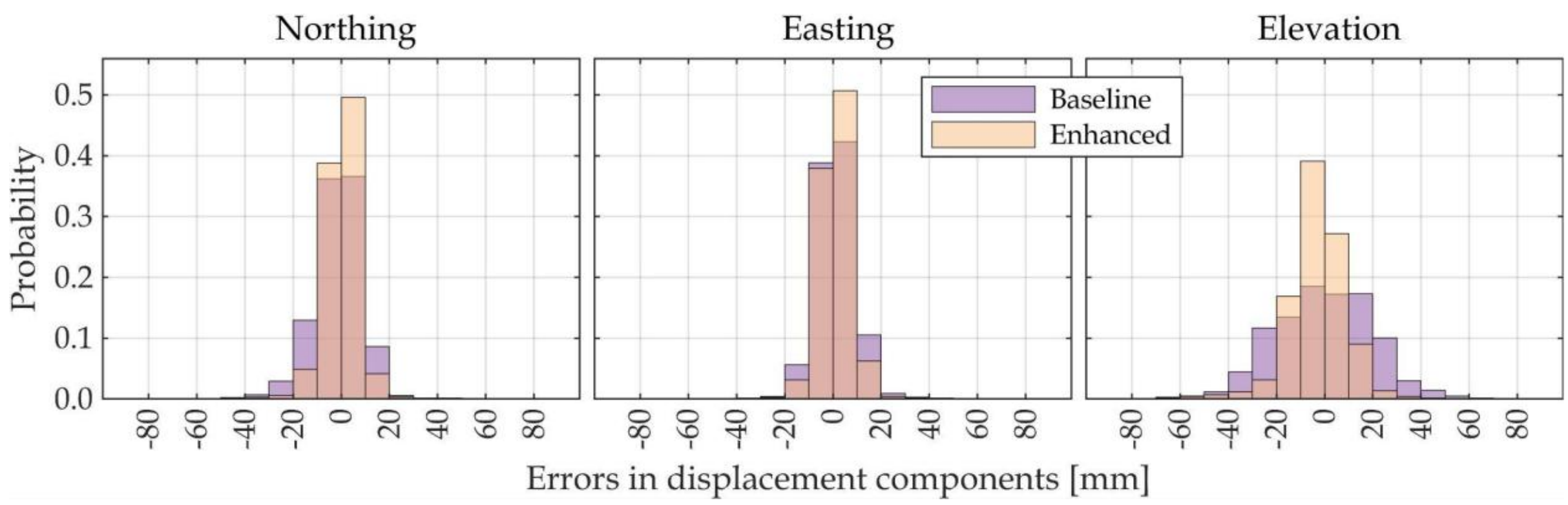

Fig. 12. Histograms of the **systematic** displacement component errors determined for the JERZ area, based on rasters generated with two different sets of processing parameters. The figure shows the distribution of the displacement component errors determined cumulatively for all analyzed periods.

This analysis suggests that for optimally chosen processing parameters (the enhanced variant), the use of sub-pixel methods for displacement analysis may be justified, even with such high-resolution products. Conversely, for the baseline variant, the large influence of systematic errors means that applying sub-pixel methods would not yield any practically significant increase in the accuracy of the final displacement measurements.

Since tilt changes and horizontal strains are derivatives of displacements, the influence of the slowly varying systematic error components on these indices is significantly less pronounced than it is for the displacements themselves (see Tables 6 and 7, and Figure 13). This is because first-difference measures naturally suppress low-frequency systematic errors. As a result, in the analyzed cases, the optimal selection of processing parameters had no practically significant impact on the accuracy of the calculated tilt change and horizontal strain values. However, it is essential to note that this does not imply that parameter selection and product quality do not affect the accuracy of these derivative indices; rather, their influence is less critical than it is for displacement determination.

Tab. 6. Summary of the accuracy assessment of tilt changes determined for the JERZ area, based on rasters generated with two different sets of processing parameters. The table provides the RMSE, the number of analyzed tilt change components, and the percentage of outliers, determined cumulatively for all analyzed periods.

| **Variant** | **Number of tilt change components** | | **Outlier percentage [%]** | **RMSE [mm/m]** |
|---|---|---|---|---|
| | $t_n$ | $t_e$ | | |
| Baseline | 4901 | 4953 | 0.0 | 0.31 |
| Enhanced | 4910 | 4963 | 0.0 | 0.27 |

Tab. 7. Summary of the accuracy assessment of horizontal strains determined for the JERZ area, based on rasters generated with two different sets of processing parameters. The table provides the RMSE, the number of analyzed horizontal strain components, and the percentage of outliers, determined cumulatively for all analyzed periods.

| **Variant** | **Number of horizontal strain components** | | **Outlier percentage [%]** | **RMSE [mm/m]** |
|---|---|---|---|---|
| | $\varepsilon_n$ | $\varepsilon_e$ | | |
| Baseline | 5169 | 5260 | 0.1 | 0.19 |
| Enhanced | 5198 | 5286 | 0.0 | 0.19 |

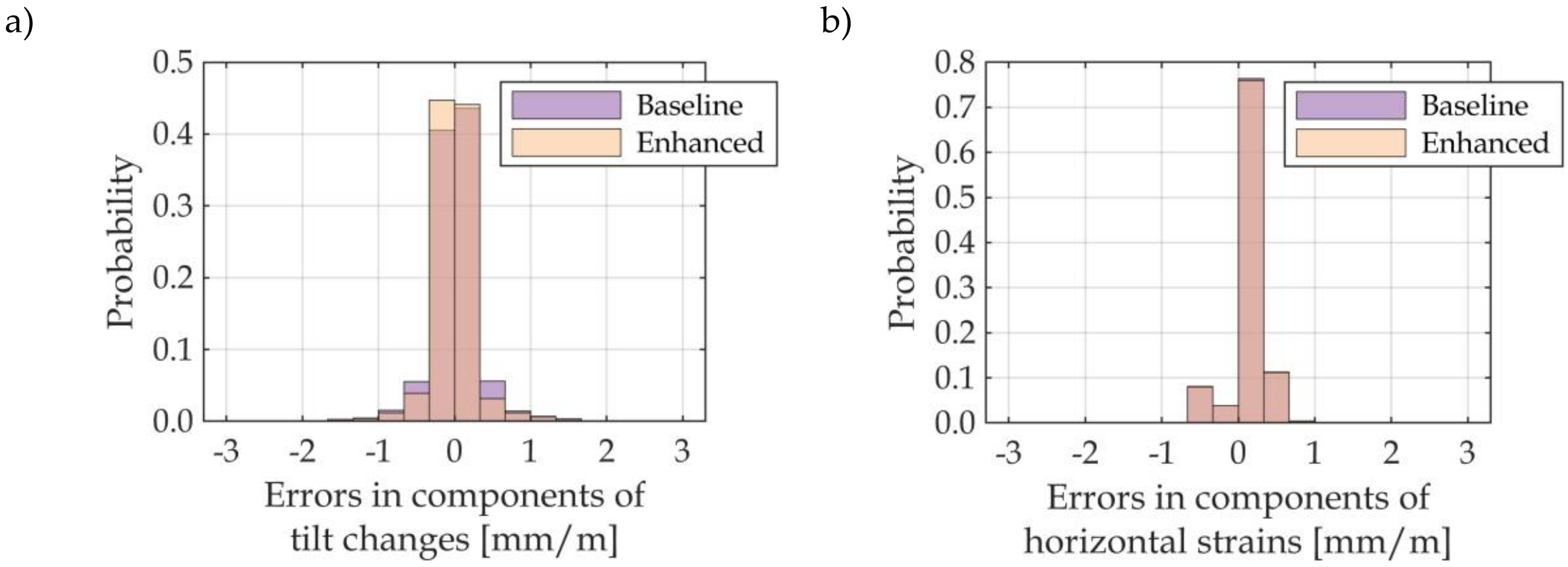

Fig. 13. Histograms of the tilt change component errors (a) and horizontal strain component errors (b) determined for the JERZ area, based on rasters generated with two different variants. The figure shows the distribution of these errors cumulatively for all analyzed periods and both components of tilt changes ($t_e$ and $t_n$) and horizontal strains ($\varepsilon_e$ and $\varepsilon_n$).

## 4. Discussion

The paper presents a thorough analysis of how processing parameters affect UAV photogrammetry product accuracy. Prior studies in this field concentrated mostly on GCP and flight configuration, thereby neglecting to analyze the influence of a considerable number of other parameters. Specifically, previous analyses have mainly examined the impact of the number and distribution of GCPs, and the effect of using GNSS RTK/RTN/PPK to determine coordinates of image projection centers in contrast to an autonomous GNSS mode. While the influence of the Brown's camera calibration model parameters has also been investigated to a certain degree, the impact of implementing the additional corrections proposed by Tang et al. [37] over the standard Brown's model [34] has not been addressed. The factors analyzed in this study also include the application of a post-BBA transformation, the *a priori* accuracy of image projection center coordinates, the assumed accuracy of tie point determination, the effect of errors assumed for GCPs, and the use of tie point filtration. These findings are especially significant for applications that demand high accuracy over large areas—a task for which a small GSD is a prerequisite, and minimization of time-consuming fieldwork is desirable. This combination of scale and resolution is a notable advance, as comparable research has typically been restricted to smaller sites (<10 ha) with coarser GSDs (20-30 mm/px).

The study evaluated the impact of the number of GCPs on the achieved accuracy. Under the tested conditions—a 200 ha study area, a GSD of 10 mm, and high-quality, well-distributed GCPs—at least five GCPs are required, which appears to be a reasonable minimum for smaller sites as well. The results indicate that for areas of ~200 ha (or larger), maintaining a GCP density of approximately 1 GCP per 40 ha substantially enhances the processing accuracy, playing a critical role for the vertical axis. Conversely, further increasing the number of GCPs (to 1 GCP per ~25 ha or 1 GCP per ~10 ha) has a relatively minor effect on improving vertical accuracy—and a negligible effect on horizontal accuracy—assuming the other processing parameters are chosen appropriately. This means that the GCP density, which can be considered sufficient, is lower than in other research concerning large areas [10, 14]. The use of a coarser GSD (17 mm and 27 mm) and smaller sensors (a Sony a6000 and the DJI Phantom 4 RTK's integrated camera) in respective studies likely accounts for the slight differences in findings presented in this paper. Importantly, GCPs also serve a critical control function. Although using more GCPs may therefore be advantageous for quality assurance, employing fewer than the recommended minimum is not advisable.

This study confirms that for high-accuracy measurements in UAV photogrammetry, a GNSS PPK workflow utilizing a local base station is superior to an RTN-based approach for image projection center determination, especially for the vertical component. While this superior performance was observed under highly controlled conditions—namely, a short baseline (<1 km for PPK vs. ~20 km to nearest CORS for RTN), a fixed and stable base station with coordinates determined based on a relatively long GNSS static session, and a unified reference frame for all GCPs and CPs—the underlying advantages of PPK are broader.

The fundamental strength of the PPK GNSS extends beyond the geometric benefit of a short baseline, which has already been proven by McMahon et al. [21]. Its inherent reliability, stemming from the elimination of continuous data link requirements and potential re-initializations, provides a more stable solution. Critically, referencing both the UAV trajectory and GCPs to the same local base station ensures exceptional spatial data consistency. This set of multifaceted advantages—stability, consistency, and the established benefit of base station proximity—reinforces the recommendation to use a PPK GNSS with a local base station for any application where accuracy is paramount. The results also show that the accuracies of the camera projection center coordinates obtained from the GNSS receiver or post-processing are more representative of the true uncertainties than uniformly assumed *a priori* values. Therefore, as a final recommendation, the *a priori* accuracies assigned to the image projection centers should reflect the real-world estimates from the GNSS post-processing.

Analysis of the impact of applying the additional corrections in the camera self-calibration process, proposed by Tang et al. [37], clearly indicates a significant improvement in accuracy compared to the utilization of only the standard Brown's model [34]. Similar to the previous cases, the use of additional corrections has the greatest influence on the accuracy of elevation determination. The impact on the accuracy of horizontal coordinates is smaller, although when using only a single GCP, the application of additional corrections can be of practical significance—especially for determining the position in the direction perpendicular to the flight path (ACR).

The conducted research also indicates that applying a seven-parameter transformation (3 translations, 3 rotations, and a scale factor) after the BBA is performed has some significance and can generally slightly improve the accuracy of position determination in the ACR direction and for the elevation. This improvement likely reflects a small offset between the coordinate system in which the GCP coordinates were determined and the system associated with the image projection centers, potentially arising from differences in the timing of GNSS observation acquisition. Adopting a tie point accuracy of 2 px, compared to the standardly used 1 px, is also of some importance for the positional accuracy in these directions.

Simultaneously, the study demonstrated that for the specific measurement technique and conditions applied, setting the *a priori* GCP coordinate accuracy in the BBA to a value within the 15 mm to 25 mm range has no practical effect on the final processing accuracy. Likewise, the research findings unequivocally point to the lack of any significant impact from applying tie point filtration. This result is noteworthy, as the filtration process involves an additional investment of time and work, yet it is a recommended step in the guidelines of the U.S. Geological Survey Open-File Report [43].

Finally, the study indicates that the optimal choice of processing parameters affects the accuracy of displacements derived from the UAV photogrammetry products. However, this impact is significantly less pronounced for deformation indices that are derivatives of displacement, such as changes in tilts and horizontal strains. This reduced importance of parameter optimization stems from the nature of the errors. Our analyses unequivocally show a spatial correlation in the errors of adjacent points. In this context, the errors are largely systematic, and this systematic component changes gradually and slowly across the spatial domain. As was the case when analyzing parameter influence via errors on CPs, the optimal parameter selection for displacement analysis is also more critical for the vertical component than for the horizontal components.

It should be noted that the results of this study are based on a single, relatively large study area and ten photogrammetric datasets collected under varying weather and vegetation conditions, all with a GSD of 10 mm. This supports the robustness of the findings, but the conclusions may not fully generalize to workflows using smaller sensors, coarser GSDs, or study sites of substantially different size. In such cases, different error characteristics and georeferencing sensitivities may emerge. The GCP coordinates used in this study were well-distributed and determined with high accuracy (using static GNSS or RTK GNSS with an on-site base station). Therefore, caution is advised when applying these conclusions to other UAV photogrammetry acquisition setups.

## 5. Conclusions

The analyses conducted indicate that the number of GCPs plays a critical role in ensuring the accuracy of results, particularly when processing parameters are poorly (randomly) selected. In such cases, a high density of GCPs acts as a fail-safe, ensuring accurate results even with suboptimal parameter selection. However, careful optimization of processing parameters can yield high accuracy with a significantly reduced number of GCPs, thereby minimizing time-consuming fieldwork.

Based on the analyses performed, the influence of individual tested parameters on accuracy can be ranked, leading to the following recommendations for similar measurement conditions (in particular, similar GSD and surveyed area):

1. Number of GCPs: While a higher number is generally beneficial, at least five well-distributed and high-quality GCPs are recommended for reliable results (NGCP ≥ 5). For areas larger than 200 ha, the number of GCPs should reflect the density of 1 GCP per 40 ha.
2. Additional corrections: These should be applied (AC = ON).
3. Method of camera projection center coordinates determination: GNSS PPK referenced to local base station is recommended (PCMD = PPK).

4. Transformation: This should be applied following the adjustment process (TR = YES).
5. *A priori* camera projection center coordinate accuracy: This should be estimated based on GNSS data post-processing (PCA = estimated).
6. Tie point accuracy: This parameter should be set to 2 pixels (TPA = 2 px).
7. *A priori* GCP coordinate accuracy: This should be set in the range between 15 mm and 25 mm (15 mm ≤ GCPA ≤ 25 mm).
8. Tie point filtration: This parameter was found to have no significant impact on results (TPF = optional).

It is important to note that while the meticulous selection of processing parameters significantly affects the accuracy of determined displacements, its influence on the accuracy of tilt change and horizontal strain is less pronounced. This discrepancy arises from the systematic nature of errors, which manifest as gradual spatial deformations in photogrammetric products. These systematic errors predominantly affect the determination of vertical displacements, leading to spatially correlated error patterns among neighboring points. In contrast, their impact on horizontal displacements is comparatively limited.

**CRediT authorship contribution statement:**

**Paweł Ćwiąkała:** Conceptualization, Methodology, Software, Validation, Formal analysis, Investigation, Resources, Data Curation, Supervision, Project administration, Funding acquisition; **Edyta Puniach:** Methodology, Software, Formal analysis, Investigation, Writing - Original Draft, Writing - Review & Editing, Visualization, Funding acquisition; **Elżbieta Pastucha:** Methodology, Software, Formal analysis, Visualization; **Wojciech Gruszczyński:** Methodology, Software, Formal analysis, Investigation, Writing - Original Draft, Writing - Review & Editing, Funding acquisition.

**Fundings:**

Research project supported by the program „Excellence initiative – research university" for the AGH University of Krakow, and statutory research of AGH University of Krakow, Faculty of Geo-Data Science, Geodesy, and Environmental Engineering [grant number 16.16.150.545].

**Declaration of Generative AI and AI-assisted technologies in the writing process**

During the preparation of this work, the authors used Gemini and ChatGPT in order to refine language. After using this tool/service, the authors reviewed and edited the content as needed and take full responsibility for the content of the publication.

**Declaration of competing interest**

The authors declare that they have no known competing financial interests or personal relationships that could have appeared to influence the work reported in this paper.

**Appendix A. Supplementary material**

Supplementary data to this article can be found online at https://doi.org/10.1016/j.measurement.2026.120315.

**Data availability**

Data will be made available on request.

**References**

[1] Pargieła, K. Rzonca, A. (2021). Determining optimal photogrammetric adjustment of images obtained from a fixed-wing UAV. The Photogrammetric Record, 36, 285-302. https://doi.org/10.1111/phor.12377

[2] Dai, W., Zheng, G., Antoniazza, G., Zhao, F., Chen, K., Lu, W. et al. (2023a) Improving UAV-SfM photogrammetry for modelling high-relief terrain: Image collection strategies and ground

control quantity. Earth Surface Processes and Landforms, 48(14), 2884–2899. https://doi.org/10.1002/esp.5665

[3] Dai, W., Qiu, R., Wang, B., Lu, W., Zheng, G., Amankwah, S. O. Y., Wang, G. (2023b). Enhancing UAV-SfM Photogrammetry for Terrain Modeling from the Perspective of Spatial Structure of Errors. Remote Sensing, 15(17), 4305. https://doi.org/10.3390/rs15174305

[4] Mueller, M. M., Dietenberger, S., Nestler, M., Hese, S., Ziemer, J., Bachmann, F., Leiber, J., Dubois, C., & Thiel, C. (2023). Novel UAV Flight Designs for Accuracy Optimization of Structure from Motion Data Products. Remote Sensing, 15(17), 4308. https://doi.org/10.3390/rs15174308

[5] Parmehr, E.G., Savadkouhi, M., Nopour, M. (2023) The impact of oblique images and flight-planning scenarios on the accuracy of UAV 3D mapping. The Photogrammetric Record, 38, 563–580. https://doi.org/10.1111/phor.12466

[6] Agüera-Vega, F., Ferrer-González, E., Martínez-Carricondo, P., Sánchez-Hermosilla, J., Carvajal-Ramírez, F. (2024). Influence of the Inclusion of Off-Nadir Images on UAV-Photogrammetry Projects from Nadir Images and AGL (Above Ground Level) or AMSL (Above Mean Sea Level) Flights. Drones, 8(11), 662. https://doi.org/10.3390/drones8110662

[7] Forlani, G., Diotri, F., Morra di Cella, U.,Roncella, R. (2020). UAV block georeferencing and control by on-board GNSS data. The International Archives of the Photogrammetry, Remote Sensing and Spatial Information Sciences, XLIII-B2-2020, 9-16. https://doi.org/10.5194/isprs-archives-XLIII-B2-2020-9-2020

[8] Zeybek, M. (2021). Accuracy assessment of direct georeferencing UAV images with onboard global navigation satellite system and comparison of CORS/RTK surveying methods. Measurement Science and Technology, 32(6), 065402. 10.1088/1361-6501/abf25d

[9] Žabota, B., Kobal, M. (2021). Accuracy Assessment of UAV-Photogrammetric-Derived Products Using PPK and GCPs in Challenging Terrains: In Search of Optimized Rockfall Mapping. Remote Sensing, 13(19), 3812. https://doi.org/10.3390/rs13193812

[10] Liu, X., Lian, X., Yang, W., Wang, F., Han, Y., Zhang, Y. (2022). Accuracy assessment of a UAV direct georeferencing method and impact of the configuration of ground control points. Drones, 6(2), 30

[11] Nesbit, P. R., Hubbard, S. M., & Hugenholtz, C. H. (2022). Direct Georeferencing UAV-SfM in High-Relief Topography: Accuracy Assessment and Alternative Ground Control Strategies along Steep Inaccessible Rock Slopes. Remote Sensing, 14(3), 490. https://doi.org/10.3390/rs14030490

[12] Cho, J. M., Lee, B. K. (2023). GCP and PPK utilization plan to deal with RTK signal interruption in RTK-UAV photogrammetry. Drones, 7(4), 265.

[13] Martínez-Carricondo, P., Agüera-Vega, F., & Carvajal-Ramírez, F. (2023). Accuracy assessment of RTK/PPK UAV-photogrammetry projects using differential corrections from multiple GNSS fixed base stations. Geocarto International, 38(1). https://doi.org/10.1080/10106049.2023.2197507

[14] Seo, D.-M., Woo, H.-J., Hong, W.-H., Seo, H., & Na, W.-J. (2024). Optimization of Number of GCPs and Placement Strategy for UAV-Based Orthophoto Production. Applied Sciences, 14(8), 3163. https://doi.org/10.3390/app14083163

[15] Miron, C.-O., Oniga, V.-E., Loghin, A.-M., Nistor, S., Chirila, C., Lazar, A.-A., Macovei, M. (2024). An In-Depth UAS-PPK Georeferencing Analysis at Different Flight Heights for an Urban Area. IEEE Access, 12, 196533-196552. doi: 10.1109/ACCESS.2024.3521991

[16] Monjardín-Armenta, S., Rangel-Peraza, J., Sanhouse-García, A., Plata-Rocha, W., Rentería-Guevara, S. & Mora-Félix, Z. (2024). Statistical comparison analysis of different real-time kinematic methods for the development of photogrammetric products: CORS-RTK, CORS-RTK + PPK, RTK-DRTK2, and RTK + DRTK2 + GCP. Open Geosciences, 16(1), 20220650. https://doi.org/10.1515/geo-2022-0650

[17] Tan, C., Chen, Z., Chen, Z., Liao, A., Zeng, X., Cao, J. (2024). Accuracy analysis of UAV aerial photogrammetry based on RTK mode, flight altitude, and number of GCPs. Measurement Science and Technology, 35(10), 106310. https://doi.org/10.1088/1361-6501/ad5dd7

[18] Sanz-Ablanedo, E., Chandler, J. H., Rodríguez-Pérez, J. R., Ordóñez, C. (2018). Accuracy of unmanned aerial vehicle (UAV) and SfM photogrammetry survey as a function of the number and location of ground control points used. Remote Sensing, 10(10), 1606.

[19] Oniga, V.-E., Breaban, A.-I., Pfeifer N., Chirila, C. (2020). Determining the suitable number of ground control points for UAS images georeferencing by varying number and spatial distribution. Remote Sensing, 12(5), 876.

[20] Zimmerman, T., Jansen, K., Miller, J. (2020). Analysis of UAS Flight Altitude and Ground Control Point Parameters on DEM Accuracy along a Complex, Developed Coastline. Remote Sensing, 12(14), 2305. https://doi.org/10.3390/rs12142305

[21] McMahon, C., Mora, O. E., & Starek, M. J. (2021). Evaluating the Performance of sUAS Photogrammetry with PPK Positioning for Infrastructure Mapping. Drones, 5(2), 50. https://doi.org/10.3390/drones5020050

[22] Arkalı, M.; Atik, M. E.: (2025). Accuracy Assessment of RTK, PPK, and PPP-AR Techniques for Direct Georeferencing in UAV-Based Photogrammetric Mapping. Int. Arch. Photogramm. Remote Sens. Spatial Inf. Sci., XLVIII-M-6-2025, 325–330. https://doi.org/10.5194/isprs-archives-XLVIII-M-6-2025-325-2025

[23] Štroner, M., Urban, R., Seidl, J., Reindl, T., Brouček, J. (2021). Photogrammetry Using UAV-Mounted GNSS RTK: Georeferencing Strategies without GCPs. Remote Sens., 13, 1336. https://doi.org/10.3390/rs13071336

[24] Dragomir, L.O., Popescu, C.A., Herbei, M.V., Popescu, G., Herbei, R.C., Salagean, T., Bruma, S., Sabou, C., Sestras, P. (2025). Enhancing Conventional Land Surveying for Cadastral Documentation in Romania with UAV Photogrammetry and SLAM. Remote Sensing, 17(13), 2113. https://doi.org/10.3390/rs17132113

[25] Stempfhuber, W., Buchholz, M. (2012). A precise low-cost RTK GNSS system for UAV applications. Int. Arch. Photogramm. Remote Sens. Spatial Inf. Sci., 1, 289-293.

[26] Turner, D., Lucieer, A., Watson, C. (2012). An automated technique for generating georectified mosaics from ultra-high resolution unmanned aerial vehicle (UAV) imagery based on structure from motion (SfM) point clouds. Remote Sens., 4(5), 1392-1410.

[27] Cledat, E., Jospin, L. V., Cucci, D. A., Skaloud, J. (2020). Mapping quality prediction for RTK/PPK-equipped micro-drones operating in complex natural environment", ISPRS J. Photogramm. Remote Sens., 167, 24-38.

[28] Zhang, H., Aldana-Jague, E., Clapuyt, F., Wilken, F., Vanacker, V., Van Oost, K. (2019). Evaluating the potential of post-processing kinematic (PPK) georeferencing for UAV-based structure- from-motion (SfM) photogrammetry and surface change detection. Earth Surf. Dyn., 7(3), 807-827.

[29] Ekaso, D., Nex, F., Kerle, N. (2020). Accuracy assessment of real-time kinematics (RTK) measurements on unmanned aerial vehicles (UAV) for direct geo-referencing". Geo-Spatial Inf. Sci., 23(2), 165-181.

[30] Sestras, P., Badea, G., Badea, A.C., Salagean, T., Roșca, S., Kader, S., Remondino, F. (2025). Land surveying with UAV photogrammetry and LiDAR for optimal building planning. Automation in Construction, 173, 106092. https://doi.org/10.1016/j.autcon.2025.106092

[31] Fraser, C.S. (1997). Digital camera self-calibration. ISPRS Journal of Photogrammetry and Remote Sensing, 52(4), 149-159. https://doi.org/10.1016/S0924-2716(97)00005-1

[32] Zhou, Y., Rupnik, E., Meynard, C., Thom, C., Pierrot-Deseilligny, M. (2019). Simulation and analysis of photogrammetric UAV image blocks: influence of camera calibration error. ISPRS

Annals of the Photogrammetry, Remote Sensing and Spatial Information Sciences, IV-2/W5, 195-200. https://doi.org/10.5194/isprs-annals-IV-2-W5-195-2019
[33] Roncella, R., & Forlani, G. (2021). UAV Block Geometry Design and Camera Calibration: A Simulation Study. Sensors, 21(18), 6090. https://doi.org/10.3390/s21186090
[34] Brown, D. C. (1971). Close-Range Camera Calibration. Photogrammetric Engineering, 37(8), 855-866.
[35] Abeho, D. R., Shoko, M., & Odera, P. A. (2024). Effects of camera calibration on the accuracy of Unmanned Aerial Vehicle sensor products. International Journal of Engineering and Geosciences, 9(3), 314-323. https://doi.org/10.26833/ijeg.1422619
[36] Wang, Z., Shi, L., Li, J., Dai, W., Lu, W., & Li, M. (2025). The Synergistic Effects of GCPs and Camera Calibration Models on UAV-SfM Photogrammetry. Drones, 9(5), 343. https://doi.org/10.3390/drones9050343
[37] Tang, R.; Fritsch, D.; Cramer, M. (2012). New rigorous and flexible Fourier self-calibration models for airborne camera calibration. ISPRS Journal of Photogrammetry and Remote Sensing, 71, 76-85. https://doi.org/10.1016/j.isprsjprs.2012.05.004
[38] Jouybari, A., Bagherbandi, M., & Nilfouroushan, F. (2021). Comparison of the strip- and block-wise aerial triangulation using different exterior orientation parameters weights. Journal of Spatial Science, 67(3), 377-394. https://doi.org/10.1080/14498596.2020.1871086
[39] Liebold, F., Mader, D., Sardemann, H., Eltner, A., & Maas, H.-G. (2023). A Bi-Radial Model for Lens Distortion Correction of Low-Cost UAV Cameras. Remote Sensing, 15(22), 5283. https://doi.org/10.3390/rs15225283
[40] Pitombeira, K. and Mitishita, E. (2023). Influence of on-site camera calibration with sub-block of images on the accuracy of spatial data obtained by ppk-based uas photogrammetry. Remote Sensing, 15(12), 3126. https://doi.org/10.3390/rs15123126
[41] Casella, V., Chiabrando, F., Franzini, M., Manzino, A.M. (2019). Accuracy Assessment of a Photogrammetric UAV Block by using Different Software and Adopting Diverse Processing Strategies. Proceedings of the 5th International Conference on Geographical Information Systems Theory, Applications and Management (GISTAM 2019), 77-87. 10.5220/0007710800770087
[42] Kostrzewa A., Płatek-Żak A., Banat P., Wilk Ł. (2025). Open-Source vs. Commercial Photogrammetry: Comparing Accuracy and Efficiency of OpenDroneMap and Agisoft Metashape. International Archives of the Photogrammetry, Remote Sensing and Spatial Information Sciences - ISPRS Archives, 48 (1-W4-2025), 65 - 72. 10.5194/isprs-archives-XLVIII-1-W4-2025-65-2025
[43] Over, J.R., Ritchie, A.C., Kranenburg, C.J., Brown, J.A., Buscombe, D., Noble, T., Sherwood, C.R., Warrick, J.A., Wernette, P.A. (2021). Processing coastal imagery with Agisoft Metashape Professional Edition, version 1.6—Structure from motion workflow documentation: U.S. Geological Survey Open-File Report 2021–1039. https://doi.org/10.3133/ofr20211039
[44] Puniach, E., Matwij, W., Gruszczyński, W., Ćwiąkała, P., Januchta, K. (2025). Determining ground surface deformation indices in urbanized mining areas based on UAV-photogrammetry products. Measurement, 253(A), 117431, https://doi.org/10.1016/j.measurement.2025.117431.
[45] Teo, T.-A., Fu, Y.-J., Li, K.-W., Weng, M.-C., Yang, C.-M. (2023). Comparison between image- and surface-derived displacement fields for landslide monitoring using an unmanned aerial vehicle. International Journal of Applied Earth Observation and Geoinformation, 116, 103164. https://doi.org/10.1016/j.jag.2022.103164
[46] He, H., Ming, Z., Zhang, J., Wang, L., Yang, R., Chen, T., Zhou, F. (2024). Robust Estimation of Landslide Displacement From Multitemporal UAV Photogrammetry-Derived Point Clouds.

IEEE Journal of Selected Topics in Applied Earth Observations and Remote Sensing, 17, 6627-6641. 10.1109/JSTARS.2024.3373505
[47] Chen, B., Maurer, J., Gong, W. (2025). Applications of UAV in landslide research: a review. Landslides. https://doi.org/10.1007/s10346-025-02547-2
[48] Puniach, E., Gruszczyński, W., Ćwiąkała, P., Matwij, W. (2021). Application of UAV-based orthomosaics for determination of horizontal displacement caused by underground mining. ISPRS Journal of Photogrammetry and Remote Sensing, 174, 282-303. https://doi.org/10.1016/j.isprsjprs.2021.02.006
[49] Ling, W., Feng, X., Wang, L., Zhu, Z., Wang, S., Fu, H., Zhang, S., Zhoa, Y. (2024). Prediction method of surface subsidence induced by block caving method based on UAV oblique photogrammetry. Sci Rep 14, 24630. https://doi.org/10.1038/s41598-024-74864-w
[50] Zhu, Y., Yan, Y., Dai, A., Dai, H., Zhang, Y., Zhang, W., Wang, Z., Li, J. (2025). UAV-MSSH: A novel UAV photogrammetry-based framework for mining surface three-dimensional movement basin monitoring. Measurement, 242(B), 115944. https://doi.org/10.1016/j.measurement.2024.115944
[51] Nakata, Y., Iwasaki, K., Shimoda, S., Torita, H. (2023). Understanding microtopography changes in agricultural landscapes through precision assessments of digital surface models by the UAV-RTK-PPK method without ground control points. Smart Agricultural Technology, 5, 100269. https://doi.org/10.1016/j.atech.2023.100269
[52] Fidelus-Orzechowska, J., Ćwiąkała, P., Puniach, E., Strzyżowski, D. (2025). Transformation of slope microrelief across an unused mountain forest road – Analysis over an 8-year period (Tatra Mts, Poland). CATENA, 256, 109085. https://doi.org/10.1016/j.catena.2025.109085.
[53] Mohren, J., Schulze, M. (2024). Automated cleaning of tie point clouds following USGS guidelines in Agisoft Metashape professional (ver. 2.1.0). MethodsX, 12, 102679. https://doi.org/10.1016/j.mex.2024.102679
[54] Puniach, E., Gruszczyński, W., Ćwiąkała, P., Strząbała, K., & Pastucha, E. (2024). Recognition of Urbanized Areas in UAV-Derived Very-High-Resolution Visible-Light Imagery. Remote Sensing, 16(18), 3444. https://doi.org/10.3390/rs16183444
[55] Diamond, A. (2025). WNCC - Weighted Normalized Cross Correlation (https://www.mathworks.com/matlabcentral/fileexchange/33340-wncc-weighted-normalized-cross-correlation), MATLAB Central File Exchange. Retrieved April 17, 2025.
[56] Torr, P. H. S.; Zisserman, A. MLESAC: A New Robust Estimator with Application to Estimating Image Geometry. Computer Vision and Image Understanding 2000, 18(1), 138–156. https://doi.org/10.1006/cviu.1999.0832
[57] Wahba, G. (1990). Spline Models for Observational Data. Society For Industrial And Applied Mathematics Philadelphia, Pennsylvania